\documentclass[10pt,twocolumn,letterpaper]{article}

\usepackage{iccv}
\usepackage{times}
\usepackage{epsfig}
\usepackage{graphicx}
\usepackage{amsmath}
\usepackage{amssymb}

\iccvfinalcopy

\usepackage{booktabs}
\usepackage{enumitem}
\usepackage{lato}
\usepackage{soul}
\usepackage[scientific-notation=true]{siunitx}
\usepackage{subcaption}
\usepackage{tabularx}
\usepackage{multicol}

\newcommand{\concept}[1]{\texttt{\small\normalfont\ttfamily\fontseries{m}\selectfont #1}}

\usepackage[pagebackref=true,breaklinks=true,letterpaper=true,colorlinks,bookmarks=false]{hyperref}

\def\iccvPaperID{11123} 

\begin{document}

\title{Agile Modeling: From Concept to Classifier in Minutes}


\author{
Otilia Stretcu{\large \thanks{Equal contribution.}} \ \textsuperscript{1},
Edward Vendrow{\large\footnotemark[1]} \   \textsuperscript{1,2},
Kenji Hata{\large\footnotemark[1]} \  \textsuperscript{1},
Krishnamurthy Viswanathan\textsuperscript{1},\\
Vittorio Ferrari\textsuperscript{1},
Sasan Tavakkol\textsuperscript{1},
Wenlei Zhou\textsuperscript{1},
Aditya Avinash\textsuperscript{1},
Enming Luo\textsuperscript{1},\\
Neil Gordon Alldrin\textsuperscript{1},
MohammadHossein Bateni\textsuperscript{1},
Gabriel Berger\textsuperscript{1},
Andrew Bunner\textsuperscript{1},\\
Chun-Ta Lu\textsuperscript{1},
Javier Rey\textsuperscript{1},
Giulia DeSalvo\textsuperscript{1},
Ranjay Krishna\textsuperscript{3},
Ariel Fuxman\textsuperscript{1}
\and
{\textsuperscript{1} Google Research}, \hspace{2mm}
{\textsuperscript{2} Stanford University}, \hspace{2mm}
{\textsuperscript{3} University of Washington} \\
{Contact: {\tt\small otiliastr@google.com}, {\tt\small afuxman@google.com}}

}


\maketitle

\begin{abstract}

The application of computer vision to nuanced subjective use cases is growing.
While crowdsourcing has served the vision community well for most objective tasks (such as labeling a ``zebra''), it now falters on tasks where there is substantial subjectivity in the concept (such as identifying ``gourmet tuna'').
However, empowering any user to develop a classifier for their concept is technically difficult: users are neither machine learning experts nor have the patience to label thousands of examples.
In reaction, we introduce the problem of Agile Modeling: the process of turning any subjective visual concept into a computer vision model through a real-time user-in-the-loop interactions.
We instantiate an Agile Modeling prototype for image classification and show through a user study (N=$14$) that users can create classifiers with minimal effort under $30$ minutes. 
We compare this user driven process with the traditional crowdsourcing paradigm and find that the crowd's notion often differs from that of the user's, especially as the concepts become more subjective.
Finally, we scale our experiments with simulations of users training classifiers for ImageNet21k categories to further demonstrate the efficacy.

\end{abstract}

\vspace{-3em}
\section{Introduction}
\label{sec:intro}

Whose voices, and therefore, whose labels should an image classifier learn from? In computer vision today, the answer to this question is often left implicit in the data collection process.
Concepts are defined by researchers before curating a dataset~\cite{deng_imagenet_2009}. 
Decisions for which images constitute positive versus negative instances are conducted by majority vote of crowd workers annotating this pre-defined set of categories~\cite{lease2011quality,sheng2008get}.
An algorithm then trains on this aggregated ground truth, learning to predict labels that represent the crowd's majoritarian consensus.

\begin{figure}[t]
  \centering
   \includegraphics[width=\linewidth]{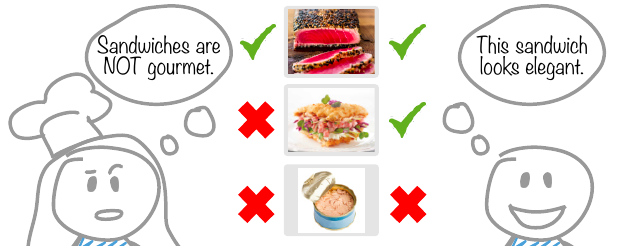}
   \caption{Visual concepts can be nuanced and subjective, differing from how a majoritarian crowd might label a concept. For example, a graduate student may think that well-prepared tuna sandwiches are considered \texttt{gourmet tuna}, but sushi chef might disagree.
   }
   \label{fig:concept-images}
    \vspace{-1em}
\end{figure}

As computer vision matures, its application to nuanced, subjective use cases is burgeoning.
While crowdsourcing has served the vision community well on many objective tasks (e.g.~identifying ImageNet~\cite{deng_imagenet_2009} concepts like ``zebra'', ``tiger''),  it now falters on tasks where there is substantial subjectivity~\cite{gordon2021disagreement}.
Everyday people want to scale their own decision-making on concepts others may find difficult to emulate: for example, in Figure~\ref{fig:concept-images}, a sushi chef might covet a classifier to source gourmet tuna for inspiration. Majority vote by crowd workers may not converge to the same definition of what makes a tuna dish gourmet.


This paper highlights the need for user-centric approaches to developing real-world classifiers for these subjective concepts.
To define this problem space, we recognize the following challenges. First, concepts are subjective, requiring users to be embedded in the data curation process. 
Second, users are usually not machine learning experts; we need interactive systems that elicit the subjective decision boundary from the user. 
Third, users don't have the patience nor resources to sift through the thousands of training instances that is typical for most image classification datasets~\cite{deng_imagenet_2009,lin2014microsoft,openimages}—for example, ImageNet annotated over 160M images to arrive at their final 14M version.

In order to tackle these challenges, we introduce the problem of \textbf{Agile Modeling}: the process of turning any visual concept into a computer vision model through a real-time user-in-the-loop process.
Just as software engineering matured from prescribed procedure to ``agile'' software packages augmenting millions of people to become software engineers, Agile Modeling aims to empower anyone to create personal, subjective vision models.
It formalizes the process by which a user can initialize and interactively guide the training process while minimizing the time and effort required to obtain a model.
With the emergent few-shot learning capabilities of vision foundation models~\cite{radford2021learning,jia2021scaling}, now is the right time to begin formalizing and developing Agile Modeling systems.

We instantiate an Agile Modeling prototype for image classification to highlight the importance of involving the user-in-the-loop when developing subjective classifiers. 
Our prototype allows users to bootstrap the learning process with a single language description of their concept (e.g.~``gourmet tuna'') by leveraging vision-language foundation models~\cite{radford2021learning,jia2021scaling}.
Next, our prototype uses active learning to identify instances that if labeled would maximally improve classifier performance. These few instances are surfaced to the user, who is only asked to identify which instances are positive, something they can do even without a background in machine learning.
This iterative process continues with more active learning steps until the user is satisfied with their classifier's performance.


Our contributions are:
\begin{enumerate}[noitemsep, topsep=0pt] 
    \item We formulate the Agile Modeling problem, which puts users at the center of the image classification process.
    \item  We demonstrate that a real-time prototype can be built by leveraging SOTA image-text co-embeddings for fast image retrieval and model training. With our optimizations, each round of active learning operates over over 10M images and can be performed on a single desktop CPU in a few minutes. In under 5 minutes, user-created models outperform zero-shot classifiers.
    \item In a setting that mimics real-world conditions, we compare models trained with labels from real users versus crowd raters. We find that the value of a user increases when the concept is nuanced or difficult.
    \item We verify the results of the user study with a simulated experiment of 100 more concepts in ImageNet21k.
\end{enumerate}

\begin{figure*}[!t]
  \centering
   \vspace{-3ex}
   \includegraphics[width=\textwidth]{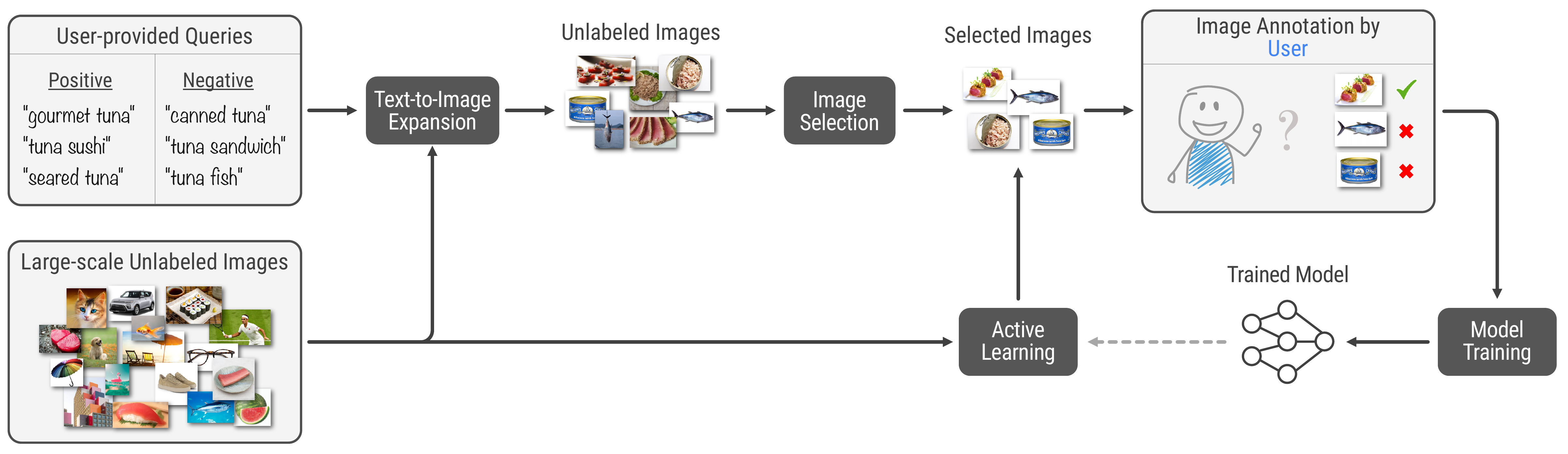}
   \vspace{-2em}
   \caption{Overview of the Agile Modeling framework. Starting with a concept in the mind of the user, the system guides the user into first defining the concept through a few text phrases, automatically expands these to small subset of images, followed by one or more rounds of real-time active learning on a large corpus, where the user only needs to rate images.
   }
   \label{fig:system}
\end{figure*}


\section{Related work}
\label{sec:related-work}
Our work draws inspiration from human-in-the-loop, personalization, few-shot, and active learning.



\vspace{1ex}
\noindent\textbf{Building models with humans-in-the-loop.}
Involving humans in the training process has a long history in crowdsourcing~\cite{fails2003interactive,amershi2014power,patel2010gestalt,fiebrink2009play}, in developmental robotics~\cite{thomaz2008teachable,knox2013learning,loftin2016learning}, and even in computer vision~\cite{krishna2022socially,daum2022vocal,zhang2023equi,krishna2021visual,park2019ai} and is recently all the rage in large language modeling~\cite{ouyang2022training}.
However, all these methods are primarily focused on improving model behavior. In other words, they ask ``how can we leverage human feedback or interactions to make a better model?'' In comparison, we take a user-centric approach and ask, ``how can we design a system that can empower users to develop models that reflect their needs?'' 

With this framing in mind, our closest related work belongs comes from the systems community~\cite{tropel, Snorkel, varma2017inferring, Mullapudi}.
Tropel~\cite{tropel} automated the process of large-scale annotation by having users provide a single positive example; and asking the crowd to determine whether other images are similar to it.
For subjective concepts, particularly those with multiple visual modes, a single image may be insufficient to convey the meaning of the concept to the crowd.
Others such as Snorkel~\cite{Snorkel, varma2017inferring} circumvented large-scale crowd labeling through the use of expert-designed labeling functions to automatically annotate a large, unlabeled dataset.
However, in computer vision, large datasets of images contain metadata that is independent of the semantics captured within the photo~\cite{thomee2016yfcc100m}.
With the recent emergent few-shot capabilities in large vision models, its now time to tackle the human-in-the-loop challenges through a modeling lens appropriate for the computer vision community.
Our prototype can train a model using active learning on millions of images on a single CPU in a matter of minutes.

\vspace{1ex}
\noindent\textbf{Personalization in computer vision.}
Although personalization~\cite{khan2021personalizing, cohen2022my, gal2022image} is an existing topic in building classification, detection, and image synthesis, these methods are often devoid of real user interactions, and test their resultant models on standard vision datasets. Conversely, we run a study with real users, focus on real-world sized datasets and on new, subjective concepts.

\vspace{1ex}
\noindent\textbf{Zero and few-shot learning.}
Since users have a limited patience for labeling, Agile Modeling aims to minimize the amount of labeling required, opting for few-shot solutions~\cite{metadataset, vinyals2016matching, shen2017deep, chen2020big, mullapudi2021background}.
Luckily, with the recent few-shot properties in vision-language models (found for example in CLIP~\cite{radford2021learning} and ALIGN~\cite{jia2021scaling}), it is now possible to bootstrap classifiers with language descriptions~\cite{pratt2022does}. 
Besides functioning as a baseline, good representations have shown to similarly bootstrap active learning~\cite{tian2020rethinking}.
We demonstrate that a few minutes of annotation by users can lead to sizeable gains over these zero-shot classifiers.

\vspace{1ex}
\noindent\textbf{Real-time active learning.}
Usually few-shot learning can only get you so far, especially for subjective concepts where a single language description or a single prototype is unlikely to capture the variance in the concept. Therefore, iterative approaches like active learning provide an appropriate formalism to maximize information about the concept while minimizing labels~\cite{al_survey,Chuang}. 
Active learning methods derive their name by ``actively'' asking users to annotate data which the model currently finds most uncertain~\cite{uncertainty} or believes is most representative of the unlabeled set~\cite{sener2017active} or both~\cite{ash2019deep,citovsky2021}.
Unfortunately, most of these methods require expensive pre-processing, reducing their utility in most real-world applications~\cite{daum2023vocalexplore}. Methods to speed up active learning limit the search for informative data points~\cite{coleman2022similarity} or use low-performing proxy models for data selection~\cite{coleman2019selection} or use heuristics~\cite{sener2017active, pinsler2019bayesian}.
We show that performing model updates and ranking images on cached co-embedding features is a scalable and effective way to conduct active learning.

\section{Agile Modeling}
\label{sec:system}

A user comes to the Agile Modeling system with just a subjective concept in mind---in our running example, \concept{gourmet tuna}. First we lay out the high level Agile Modeling problem framework and then describe how we instantiate a prototype of this framework.

\subsection{The framework}
As shown in Figure~\ref{fig:system}, the Agile Modeling framework guides the user through the creation of a image classifier through the following steps:
\setul{3pt}{.4pt}
\begin{enumerate}[noitemsep] 
    \item \ul{Concept definition.} The user describes the concept using text phrases. They are allowed to specify both positive phrases, which can describe the concept as a whole or specific visual modes, as well as negative phrases, which are related but not necessarily part of the concept (\eg \concept{canned tuna} is not gourmet).
    \item \ul{Text-to-image expansion and image selection.} The text phrases are used to mine relevant images from a large unlabeled dataset of images for the user to rate.
    \item \ul{Rating.} The user rates these images through a rating tool, specifying whether each image is either \texttt{positive} or \texttt{negative} for the concept of interest.
    \item \ul{Model training.} The rated images are used to train a binary classifier for the concept. This is handled automatically by the system.
    \item \ul{Active learning.} The initial model can be improved very quickly via one or more rounds of active learning. This consists of 3 repeated steps: (1) the framework invokes an algorithm to select from millions of unlabeled images to rate; (2) the user rates these images; (3) the system retrains the classifier with all the available labeled data. The whole active learning procedure operates on millions of images and returns a new model in under 3 minutes (measured in Section~\ref{sec:obtaining-model-in-minutes}).
\end{enumerate}

The user's input is used for only two types of tasks, which require no machine learning experience: first in providing the text phrases and second in rating images. Everything else, including data selection and model training, is performed automatically. With such an automated process, users do not need to hire an machine learning or computer vision engineer to build their classifiers.


\subsection{The prototype}
We focus our prototype on the core north star task of image classification~\cite{fei2022searching}. One of the main challenges of Agile Modeling is to enable the user to effectively transfer their subjective interpretation of a concept into an operational machine learning model. For our image classification task, Agile Modeling seeks to turn this arbitrary concept into a well-curated training dataset of images. We assume that that the user only has access to a large, unlabeled dataset, which is something that is easily available through the internet~\cite{radford2021learning}. Our aim is to select and label a small subset of this large dataset and use it as training data.

\vspace{1ex}
\noindent\textbf{Concept definition.}
Users initiate the Agile Modeling process by expressing their concept in words. For example, the user might come in and simply say \concept{gourmet tuna}.
However, users can preemptively also provide more than a single phrase. They can also produce negative descriptions of what their concept is \textit{not}. They can clarify that \concept{canned tuna} is not gourmet.
Through our interactions with users, we find that expressing the concept in terms of both positive and negative phrases is an effective way of mining positive and hard negative examples for training.
The positive phrases allow the user to express both the concept as a whole (\eg \concept{gourmet tuna}) and specific visual modes of it (\eg \concept{seared tuna}, \concept{tuna sushi}). The negative phrases are important in finding negative examples that could be easily confused by raters.

 
\vspace{1ex}
\noindent\textbf{Text-to-image expansion and image selection.}
\label{sec:query-expansion}
The phrases provided by the user are used to identify a first set of relevant training images. To achieve this, we take advantage of recent, powerful image-text models, such as CLIP~\cite{radford2021learning} and ALIGN~\cite{jia2021scaling}. We co-embed both the unlabeled image dataset and the text phrases provided by the user into the same space, and perform a nearest-neighbors search to retrieve 100 images nearest to each text embedding. We use an existing nearest-neighbors implementation~\cite{wu2017multiscale, guo2016quantization} that is extremely fast due to its hybrid solution of trees and quantization. From the set of all nearest neighbors, we randomly sample 100 images for the user to rate. We do this for both positive and negative phrases, since the negative texts are helpful in identifying hard negative examples.

\vspace{1ex}
\noindent\textbf{Data labeling by user.} 
The selected images are shown to the user for labeling. In our experiments, we created a simple user interface where the user is shown one image at a time and is asked to select whether it is \texttt{positive} or \texttt{negative}. The median time for our users to rate a single image was $1.7 \pm 0.5$ seconds. Since users rate 100 images per annotation round, they spend approximately 3 minutes before a new model is trained.

\vspace{1ex}
\noindent\textbf{Model training.}
We train our binary image classifier using all previously labeled data. This setup is challenging because there is little data available to train a generalizable model, and the entire training process must be fast to enable real-time engagement with the user waiting for the next phase of images.
The lack of large-scale data suggests the use of few-shot techniques created to tackle low data scenarios, such as meta-learning~\cite{wang2020generalizing, hochreiter2001learning, finn2017model} or prototype methods~\cite{snell2017prototypical}, however most such approaches are too slow for a real-time user interaction.
While the study of real-time few-shot methods is an interesting problem for future instantiations of the Agile Modeling framework, we adopted another solution that helps us address both challenges: we again take advantage of powerful pretrained models like CLIP and ALIGN to train a small multilayer perceptron (MLP), with only 1-3 layers, on top of image embeddings provided by such large pretrained models. These embeddings bring much needed external information to address the low data challenge while allowing us to train a low-capacity model that can be trained fast and is less prone to overfitting. Model architectures and training details are described in Section~\ref{sec:experiments}.

\vspace{1ex}
\noindent\textbf{Active learning (AL).}
\label{sec:al}
We improve the classifier in the traditional model-based active learning fashion: (1) we use the current model to run inference on a large unlabeled pool of data, (2) we carefully select a batch of images that should be useful in improving the model, (3) we rate these images, (4) we retrain the model. This process can be repeated one or more times to iteratively improve performance. 
When selecting samples to rate, state-of-the-art AL methods generally optimize for improving the model fastest~\cite{ren2021survey}. However, when the user is the rater, we have a real-time constraint to minimize the user-perceived latency. Therefore, AL methods that rely on heavy optimization strategies cannot be used. In our solution, we adopt a well-known and fast method called {\em uncertainty sampling} or {\em margin sampling}~\cite{culotta2005reducing, scheffer2001active, lewis1995sequential}, which selects images for which the model is {\em uncertain}. Specifically, given a model with parameters $\theta$ and a sample $x$, we define the uncertainty score as $ P_{\theta}(\hat y_1|x) -  P_{\theta}(\hat y_2|x)$, where $\hat y_1$  and $\hat y_2$ are the highest and second-highest probabilities predicted by the model.
Note there are other definitions of uncertainty such as least confidence and entropy, but since we are in a binary classification setting, all of these definitions are mathematically equivalent. 
We run one or more rounds of AL, the number of rounds is determined by the time the user has.
\section{Experiments with real users}
\label{sec:experiments}

We run user studies with real users in the loop, and show that:
\textbf{(1)} In only 5 minutes, the performance of an Agile model can exceed that of state-of-the-art zero-shot models based on CLIP and ALIGN by at least $3\%$ AUC PR (Section~\ref{sec:obtaining-model-in-minutes});
\textbf{(2)} For hard, nuanced concepts, Agile models trained with user annotations outperform those trained with crowd annotations even when crowd raters annotate 5x more data (Section~\ref{sec:value-of-domain-experts});
\textbf{(3)} Smaller active learning batch sizes perform better than larger ones, but there is an efficiency trade-off (Section~\ref{sec:ablation});
\textbf{(4)} Agile models using ALIGN embeddings outperform does using CLIP throughout model iterations (Section~\ref{sec:ablation});

\subsection{Choosing subjective concepts}

\vspace{1ex}
\noindent\textbf{Concepts.} For our user studies we select a list of $14$ novel concepts, spanning different degrees of ambiguity and difficulty. The list ranges from more objective concepts such as \concept{pie chart}, \concept{in-ear headphones} or \concept{single sneaker on white background}, to more subjective ones such as \concept{gourmet tuna}, \concept{healthy dish}, or \concept{home fragrance}. 
We found that our concepts cover a large spread over the visual space; we measure this spread using the average pairwise cosine distance between the concept text embeddings (using CLIP). For our $14$ concepts, the average pairwise cosine distance was $0.73 \pm 0.13$. In comparison, ImageNet's average pairwise cosine distance was $0.35 \pm 0.11$.
The full list of concepts is included in Appendix~\ref{appendix:concepts}, along with the queries provided by the users.

\vspace{1ex}
\noindent\textbf{Workflow.} We provide users with only the concept name and a brief description, but allow them to define the full interpretation. 
For instance, one of our users, who was provided with the concept \concept{stop-sign}, limited its interpretation to only real-world stop-signs: only stop signs in traffic were considered positive, while stop-sign drawings, stickers, or posters were considered negative\footnote{This definition was inspired by a self-driving car application, where a car should only react to real stop signs, not those on posters or ads.}.


\begin{table}
  \centering
  \small
  \vspace{-1ex}
  \begin{tabular}{l|r}
    \toprule
    \multicolumn{1}{c|}{Step} & \multicolumn{1}{c}{Time}  \\
    \midrule
    User rates 100 images & 2 min 49 sec $\pm$ 58 sec\\
    AL on 10M images & 58.6 sec $\pm$ 0.8 sec \\
    Training a new model & 23.1 sec $\pm$ 0.2 sec\\
    \bottomrule
  \end{tabular}
  \vspace{-1ex}
  \caption{The average and standard deviation of the time it takes per step in our Agile Modeling instantiation. Rating time was measured by taking the average median time of an user to rate one image during the experiments used in this paper. To measure time for AL and model training, they were each run 10 times.}
  \label{tab:time-per-step}
  \vspace{-1em}
\end{table}

\vspace{1ex}
\noindent\textbf{Participants.} When collecting data for the experiments, we sourced $14$ volunteer users to interact with our system. Each participant built a different concept. None of the users performed any machine learning engineering tasks. Our experiments indicate that it takes participants $2$ minutes and $49$ seconds on average to label $100$ images, as shown in Table~\ref{tab:time-per-step}. Our participants were adults that spanned a variety of age ranges (18-54), gender identities (male, female), and ethnicities (White, Asian, and Middle Eastern).

\vspace{1ex}
\noindent\textbf{Data sources.} Since our prototype requires an unlabeled source of images from which to source training labels, we use the LAION-400M dataset~\cite{schuhmann2021laion}, due to its large size and comprehensive construction based on the large Common Crawl web corpus. We throw away the text associated with the images. We remove duplicate URLs and split imags into a $100$M training and $100$M testing images. All Agile models trained use data exclusively from the unlabeled training split, including during nearest neighbor search, active learning, and training. For evaluation, we only use data from the $100$M test set, where each concept's evaluation set consists of a subset of this data rated by the user.

\subsection{Experimental setup}
\label{sec:experimental_setup}

\textbf{Models and training.}
All models are multilayer perceptrons (MLP) that take image representations from a frozen pretrained model as input and contain one or more hidden layers. For the first active learning step, we use a smaller MLP with $1$ hidden layer of $16$ units to prevent overfitting, while all active learning rounds and final model have $3$ hidden layers of size $128$. 
All training details, including optimizer, learning rate, etc., can be found in Appendix~\ref{appendix:experimental-details}.

\vspace{1ex}
\noindent\textbf{Baselines.}
One baseline we compare against is zero-shot learning, which corresponds to zero effort from the user. We implement a zero-shot baseline that scores an image by the cosine similarity between the image embedding and the text embedding of the desired concept. 
We also compare against a recently released active learning algorithm for learning rare vision categories~\cite{Mullapudi}. This system is the most relevant related work. We replace our active learning algorithm with theirs and compare the performance in Section~\ref{sec:ablation}.

\begin{figure}[t]
  \centering
  \vspace{-1ex}
   \includegraphics[width=\linewidth]{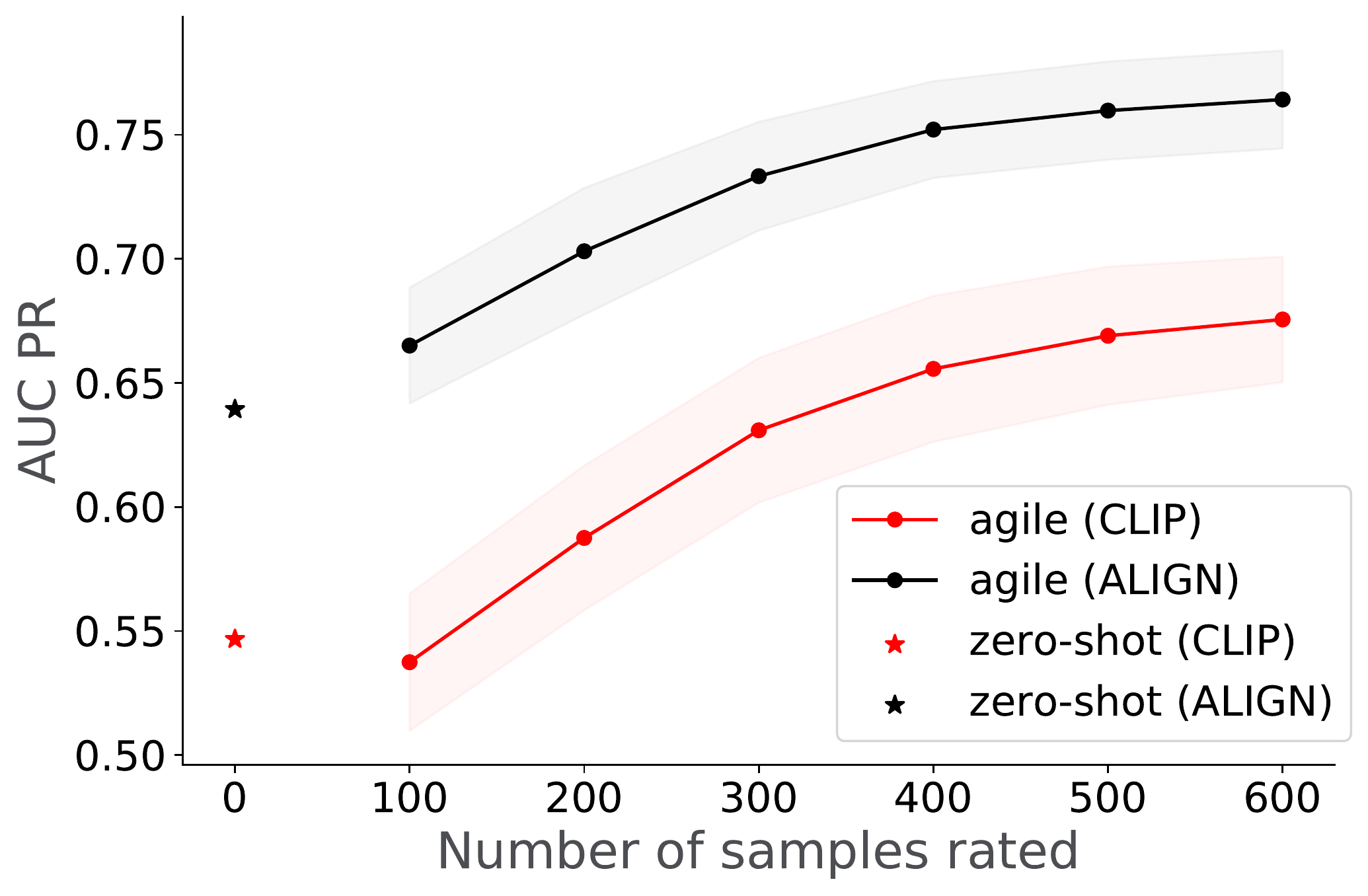}
   \vspace{-1.5em}
   \caption{Model performance per amount of samples rated by the user (AUC PR mean and standard error over all concepts). Each $\bullet$ corresponds to an active learning round.}
   \label{fig:zs-vs-base-auc-pr}
  \vspace{-1em}
\end{figure}

\begin{figure*}[t]
     \centering
     \vspace{-2ex}
     \begin{subfigure}[b]{0.33\linewidth}
         \centering
         \vspace{-1ex}
         \includegraphics[width=\linewidth]{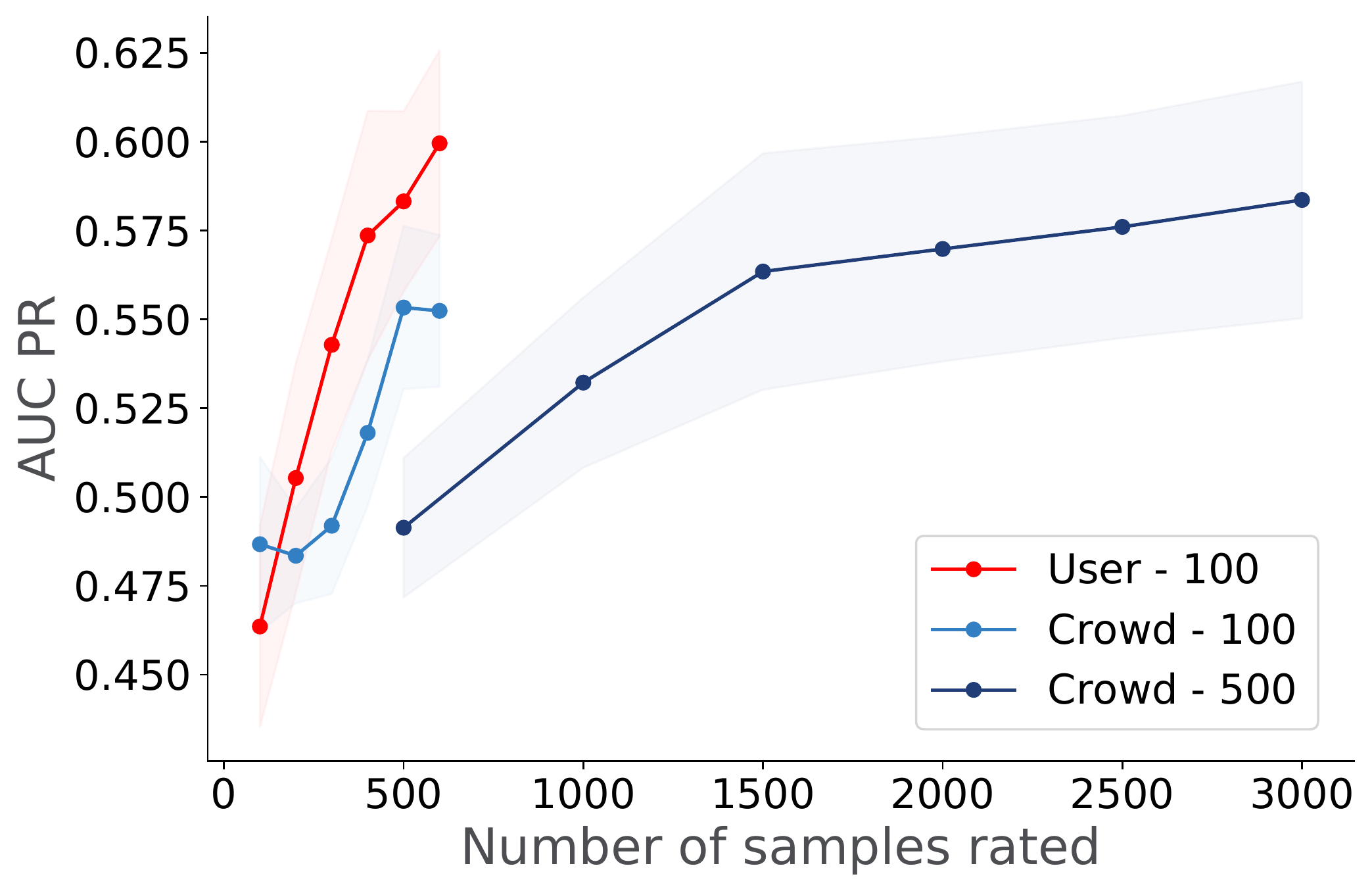}
         \caption{Hard concepts -- per \# samples rated}
         \label{fig:al-auc-pr-crowd-hardest-per-sample}
     \end{subfigure}
     \hfill
     \begin{subfigure}[b]{0.33\linewidth}
         \centering
         \vspace{-1ex}
         \includegraphics[width=\linewidth]{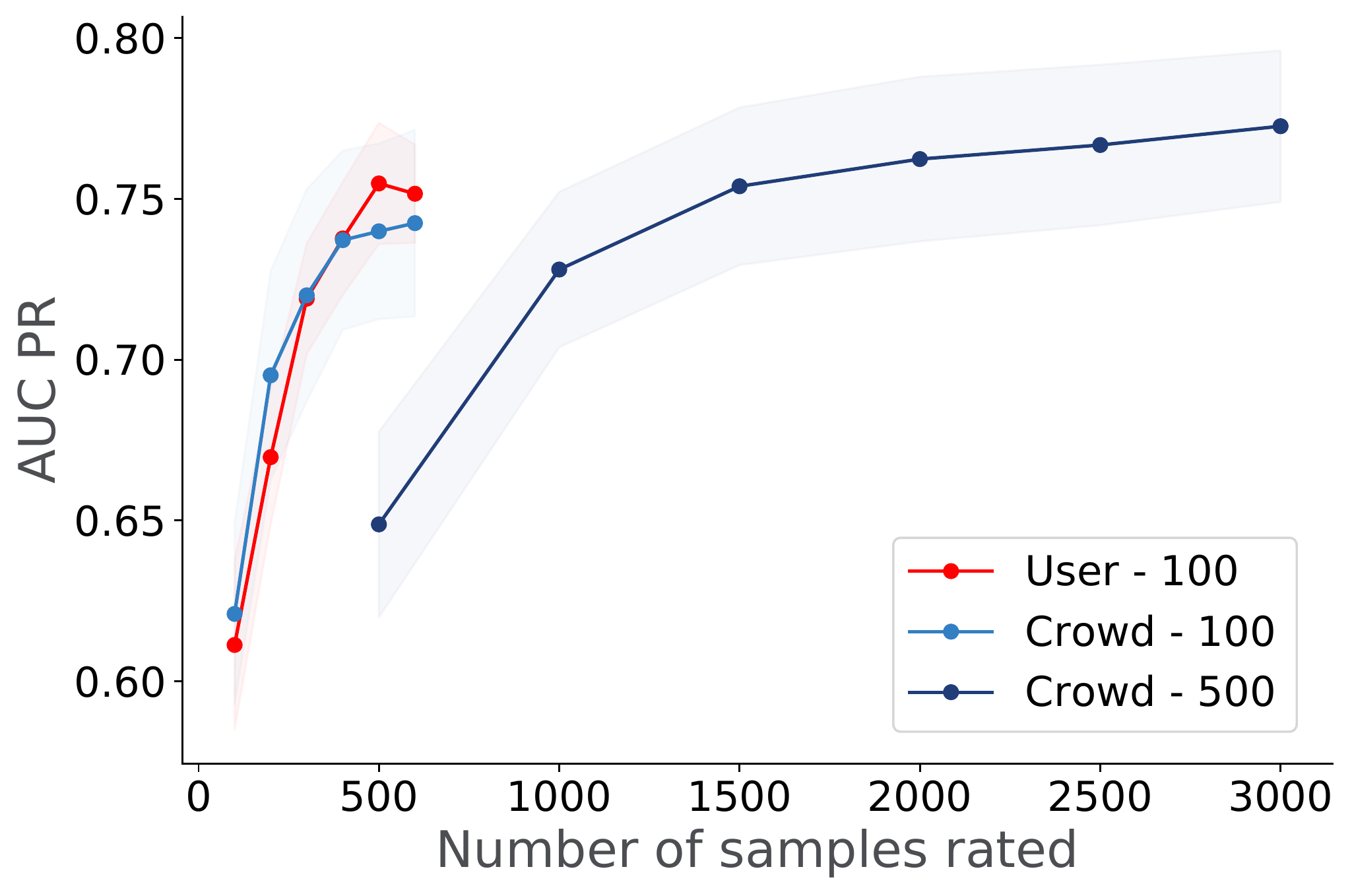}
         \caption{Easy concepts -- per \# samples rated}
         \label{fig:al-auc-pr-crowd-easiest-per-sample}
     \end{subfigure}
     \hfill
     \begin{subfigure}[b]{0.33\linewidth}
         \centering
         \vspace{-1ex}
         \includegraphics[width=\linewidth]{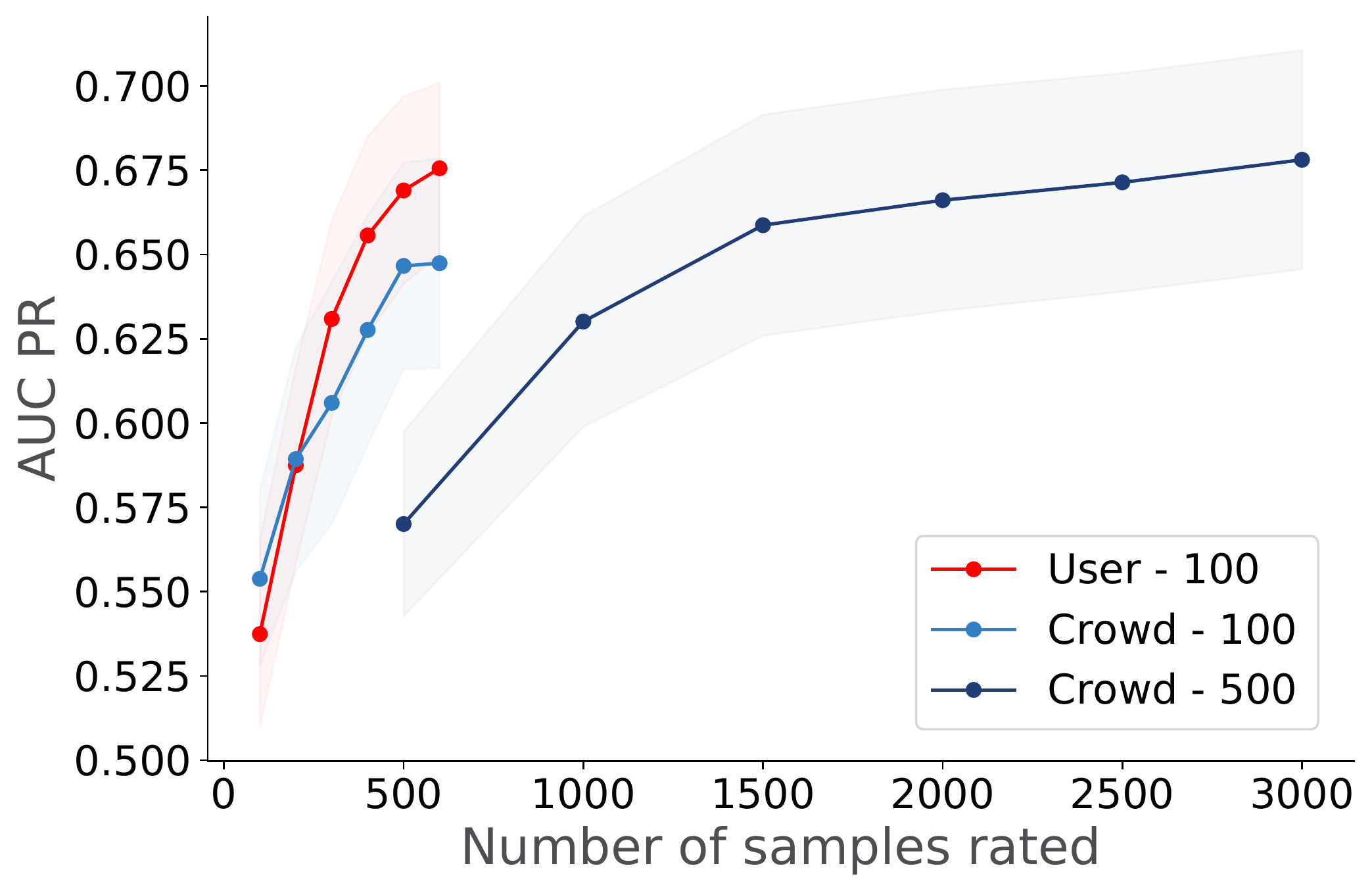}
         \caption{All concepts -- per \# samples rated}
         \label{fig:al-auc-pr-crowd-all-per-sample}
     \end{subfigure}
    
     \vspace{-1ex}
     \caption{Performance per \# samples rated by the user or crowd. AUC PR mean and standard error over subsets of concepts: hardest for the zero-shot model (left), easiest for the zero-shot model (middle), all (right). Each $\bullet$ represents an AL round.
        }
        \label{fig:al-auc-pr-crowd}
\end{figure*}

\begin{figure*}[t]
  \centering
  \vspace{-1ex}
   \includegraphics[width=\linewidth]{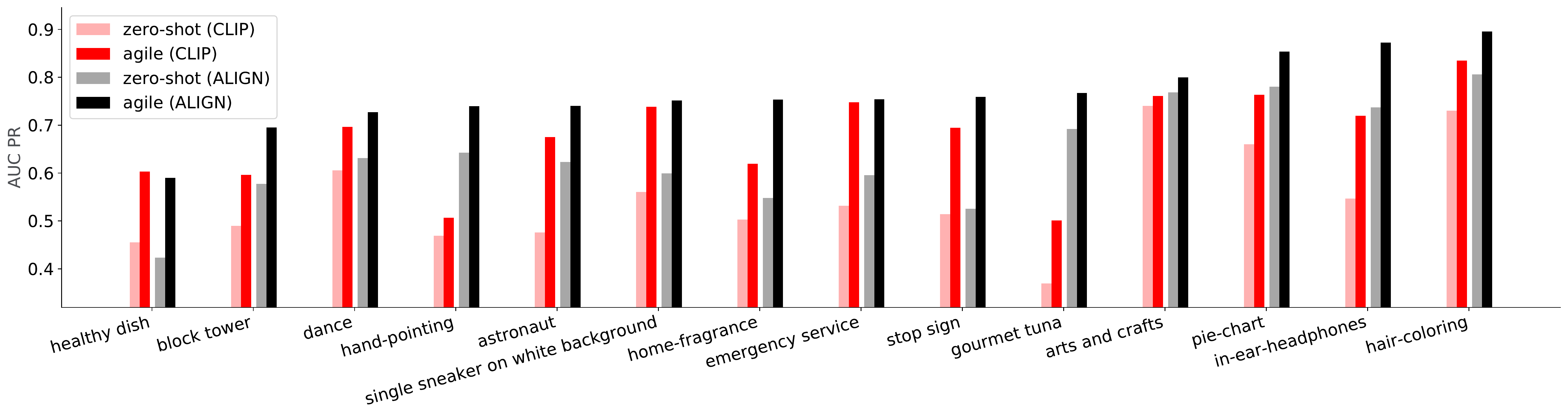}
   \vspace{-2.2em}
   \caption{Model performance per concept for zero-shot and user-in-the-loop Agile models on CLIP and ALIGN embeddings.}
     \vspace{-1ex}
   \label{fig:results_per_concept-auc-pr}
\end{figure*}

\vspace{1ex}
\noindent\textbf{Evaluation protocol.} To evaluate the models trained with the Agile Modeling prototype, we require an appropriate test set. Ideally, the user would provide a comprehensive test set---for example, ImageNet holds out a test set from their collected data~\cite{russakovsky2015imagenet}. However, since our users are volunteers with limited annotation time, they cannot feasibly label the entire LAION-400M dataset or its $100$M test split. Additionally, since we are considering rare concepts, labeling a random subset of unlabeled images is unlikely to yield enough positives.
To address these problems, we ran stratified sampling on each model, which divides images based on their model score into $10$ strata ranging from $[0,0.1)$ to $[0.9,1.0]$. In each strata, we hash each image URL to a 64-bit integer using the pseudorandom function SipHash~\cite{aumasson2012siphash} and include the 20 images with the lowest hashes in the evaluation set. Each model contributes equally to final test set. The final evaluation set has over $500$ images per category with approximately $50\%$ positive rate. The full details of the evaluation set distribution and acknowledgement of its potential biases can be found in Appendix~\ref{appendix:evaluation_strategy}.

\vspace{1ex}
\noindent\textbf{Other hyperparameters.} The text-to-image expansion expands each user-provided query to $100$ nearest-neighbor images. Next, the image selection stage randomly selects a total of $100$ images from all queries, leading to an initial training set of $100$ samples for the first model. Users are asked to perform $5$ rounds of active learning, rating $100$ images per step. These hyperparameters were chosen based on two held-out concepts, and the ablation results in Section~\ref{sec:ablation}.

\subsection{Results}

\subsubsection{Users produce classifiers in minutes}
\label{sec:obtaining-model-in-minutes}

A key value proposition of Agile Modeling is that the user should be able to train a model in minutes. We now report the feasibility of this proposition.

\vspace{1ex}
\noindent\textbf{Measuring Time.}
The time it takes per for each step of the framework is detailed in Table~\ref{tab:time-per-step}. Our proposed Agile Modeling implementation trains one initial model and conducts five active learning rounds, taking 24 minutes on average to generate a final model.

\vspace{1ex}
\noindent\textbf{Comparison with zero-shot.} 
We start by comparing against zero-shot classification, which corresponds to a scenario with minimal effort from the user. In  Figure~\ref{fig:zs-vs-base-auc-pr}, we present the performance our instantiations of the Agile Modeling framework against a zero-shot baseline across two image-text co-embeddings: CLIP~\cite{radford2021learning} and ALIGN~\cite{jia2021scaling}.
We find that the zero-shot  performance is roughly on par as a supervised model trained on 100 labeled examples by the user.
However, after the user spends a few more minutes rating (i.e., as the number of user ratings increases from 100 to 600), the resulting supervised model outperforms zero-shot. 


\vspace{1ex}
\noindent\textbf{User time versus performance.} To measure the trade-off between user time versus model performance, we show in Figure~\ref{fig:zs-vs-base-auc-pr} the AUC PR of the model across active learning rounds. We include additional metrics in Appendix~\ref{appendix:al-results}.
We include results for both CLIP and ALIGN representations as input to our classifiers. We also compare against the respective zero-shot models using CLIP and ALIGN, which are considered the zero effort case.
For both types of representations, we see a steeper increase in performance for the first 3 active learning rounds, after which the performance starts to plateau, consistent with existing literature applying active learning to computer vision tasks~\cite{karamcheti2021mind}.
Interestingly, for CLIP representations, the initial model trained on only $100$ images performs worse than the zero-shot baseline, but the zero-shot model is outperformed with just one round of active learning. We do not see this effect on ALIGN representations, where even 100 samples are enough to outperform the zero-shot model---perhaps because ALIGN representations are more effective.
We compare CLIP and ALIGN in more detail in Section~\ref{sec:ablation}.
Importantly, We show that with only 5 minutes of the user's time (Table~\ref{tab:time-per-step}), we can obtain a model that outperforms the zero-shot baseline by at least $3\%$. After 24 minutes, this performance gain grows to 16\%.

\begin{figure}[!t]
  \centering
  \vspace{-2ex}
   \includegraphics[width=0.95\linewidth]{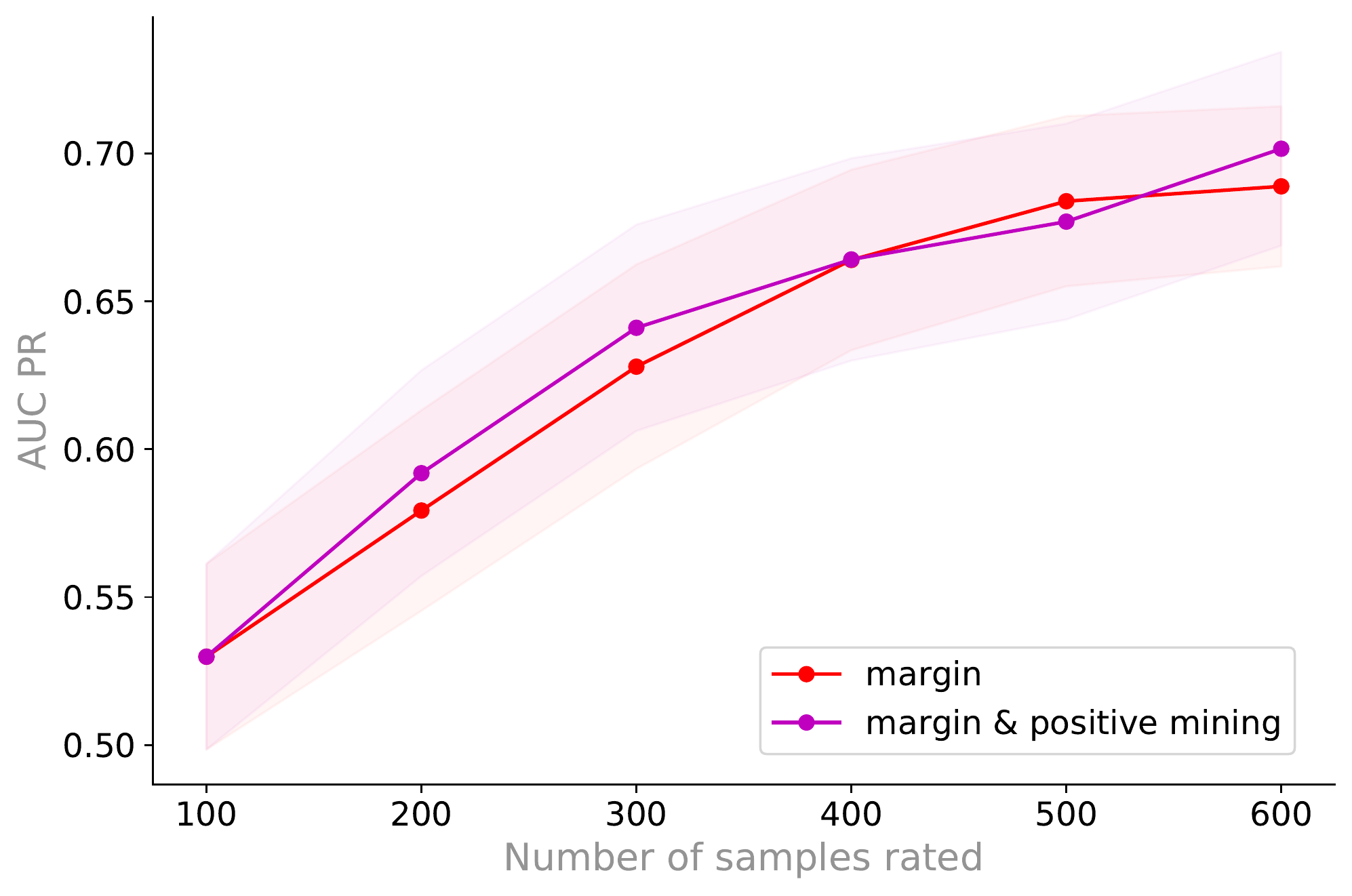}
    \vspace{-1ex}
   \caption{Model performance for two active learning methods: margin and the approach of \cite{Mullapudi} (margin \& positive mining). Each $\bullet$ corresponds to an AL round. We show the AUC PR mean and standard error over all concepts.}
   \label{fig:al_stanford}
   \vspace{-3ex}
\end{figure}

\subsubsection{Value of users in the loop versus crowd workers}
\label{sec:value-of-domain-experts}

We now study the value of empowering users to train models by themselves. In particular, we address the following question: Are there concepts for which a user-centered Agile framework leads to better performance?

Users have an advantage over crowd raters in their ability to rate images according to their subjective specifications. However, this subjectivity, or ``concept difficulty'' varies by concept: if a concept is universally understood, the advantage diminishes. Conversely, complex, nuanced concepts are harder for crowd workers to accurately label. To take this into consideration, we first partition the concepts into two datasets based on their difficulty, using zero-shot performance
as a proxy for concept difficulty. The $7$ concepts that admit the highest zero-shot performance are considered ``easy," while the remaining $7$ concepts are considered ``hard." The specific groups can be found in Appendix~\ref{appendix:concept-difficulty}. Notice that the ``difficult'' concepts include more subjective concepts such as \concept{gourmet tuna} (as illustrated in Figure~\ref{fig:concept-images}), or with multiple and ambiguous modes such as \concept{healthy dish}; whereas the ``easy'' concepts include simple, self-explanatory concepts such as \concept{dance} or \concept{single sneaker on white background}.

We then evaluate models trained by three sets of raters:
\begin{enumerate}[noitemsep, leftmargin=1em,topsep=0pt]
\item \texttt{User-100}: Users rate 100 images for the initial model and every AL round (total 600 images).
\item \texttt{Crowd-100}: Crowd workers rate 100 images for the initial model and every AL round (total 600 images).
\item \texttt{Crowd-500}: Crowd workers rate 500 images for the initial model and every AL round (total 3000 images).
\end{enumerate}

The only difference in the configurations above is who the raters are (user or crowd) and the total number of ratings. For crowd ratings, having clear instructions is crucial for accurate results, but obtaining them is a non-trivial task in the machine learning process~\cite{gadiraju2017clarity, dow2012shepherding}.
In this experiment, crowd workers read instructions created by the users, who noted difficult cases that they found during labeling. Details about the crowd instructions can be found in Appendix~\ref{appendix:crowd-task-design}. 

We plot the results in Figure~\ref{fig:al-auc-pr-crowd}, which shows the average performance for the ``hard'', ``easy'' and all concepts as a function of the number of rated samples, using CLIP embeddings. 
Per-concept results can be found in Appendix~\ref{appendix:crowd}.
On hard concepts, models trained with users (User-100) outperform models trained with crowd raters, even when $5\times$ more ratings are obtained from the crowd (Crowd-500). This suggests that Agile Modeling is particularly useful for harder, more nuanced and subjective concepts.

\subsection{Ablation studies}
\label{sec:ablation}




Although our main contribution is introducing the problem of Agile Modeling, instantiating our prototype explores a number of design decisions. In this section, we lay out how these designs change the outcome.

\begin{figure}[!t]
  \centering
  \vspace{-2ex}
   \includegraphics[width=0.95\linewidth]{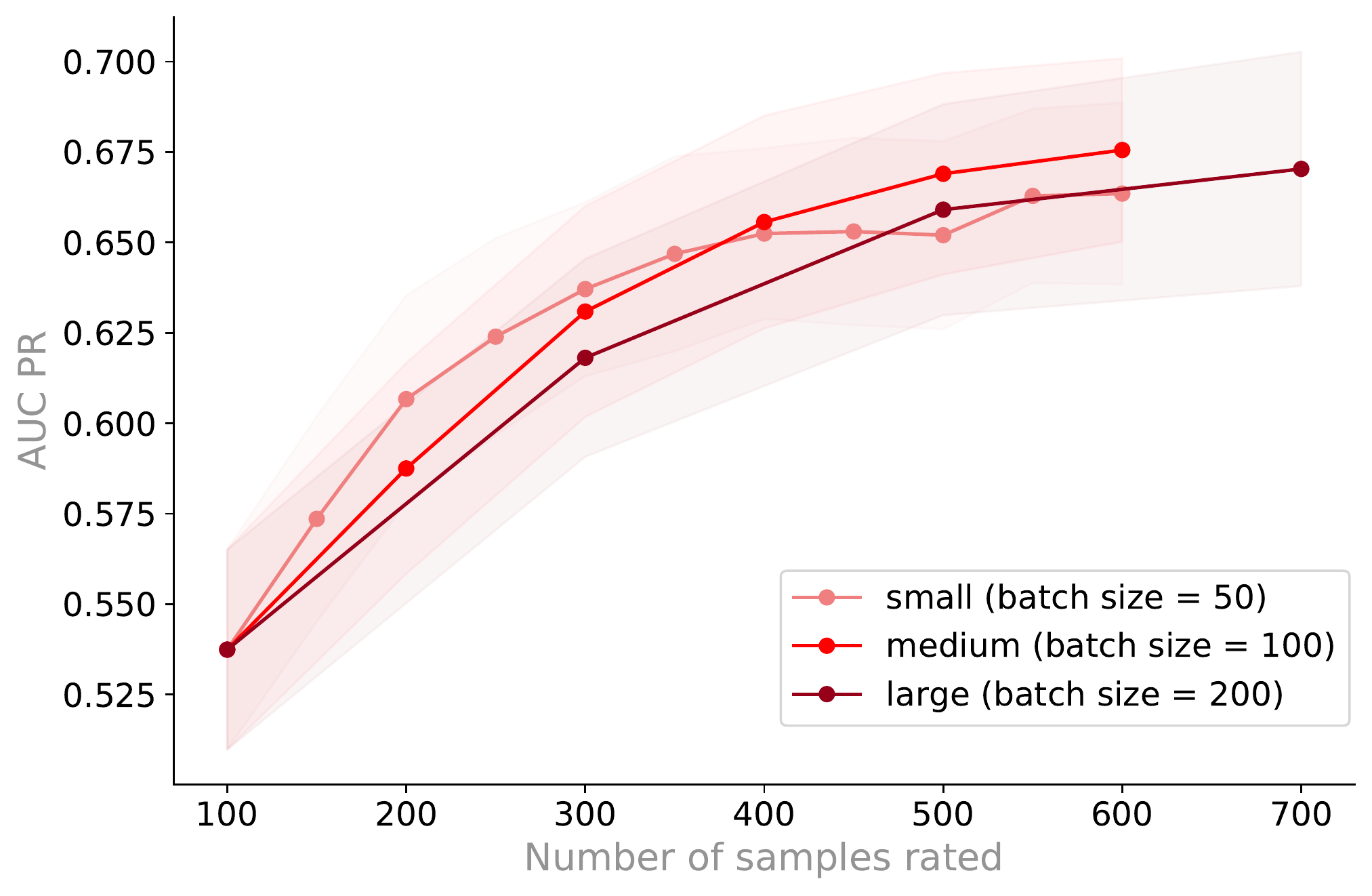}
   \vspace{-1ex}
   \caption{Model performance during active learning with 3 AL batch sizes: small (50), medium (100), large (200). Each $\bullet$ corresponds to an AL round. We show the AUC PR mean and standard error over all concepts.}
   \label{fig:batch_sizes}
   \vspace{-3ex}
\end{figure}

\vspace{1ex}
\noindent\textbf{Active learning method.}
Throughout the paper, we instantiate the active learning component with the well-known margin method~\cite{margin_sampling}. We now compare it to the active learning method used in Mullapudi et al \cite{Mullapudi}.  We ran a version of our instantiation of the Agile framework where we replace margin with the margin+positive mining strategy chosen by \cite{Mullapudi} and described in Section~\ref{sec:al}. The performance of the two methods per AL round is shown in Figure~\ref{fig:al_stanford}. Interestingly, despite the fact that Mullapudi et al. \cite{Mullapudi} introduced this hybrid approach to improve upon margin sampling, in this setting the two methods perform similarly across all AL rounds. We see the same effect on most concepts when inspecting on a per-concept basis in Appendix~\ref{appendix:al-results}.
One potential explanation for this is that the initial model trained before AL is already good enough (perhaps due to the powerful CLIP embeddings) for margin sampling to produce a dataset balanced in terms of positive and negative, and thus explicitly mining easy positives as in~\cite{Mullapudi} is not particularly useful.
Since the two methods perform equivalently, we opted for the simpler and more efficient margin in the rest of the experiments.

\vspace{1ex}
\noindent\textbf{Active learning batch size.}
Our prototype asks the user to annotate images across $5$ rounds of active learning, $100$ images per round. However, we can simultaneously change the number of images rated per round and the number of active learning rounds the user conducts. We evaluate the downstream effects of changing active learning batch size and number of rounds on model performance and time spent. We consider $3$ batch sizes: small ($50$ images/batch), medium ($100$ images/batch), large ($200$ images/batch). We run repeated rounds of active learning with each of these settings, retraining the model after each round using CLIP representations. The results in Figure~\ref{fig:batch_sizes} show that, for a fixed amount of images rated, smaller batch sizes are better than larger, especially so in the beginning. This result is expected, because for a fixed rating budget, the smaller batch setting has the chance to update the model more frequently.
While these results suggest that we should opt for a smaller batch size, there is still a trade-off between user time and performance, even when we have the same total number of samples rated. That is because model training takes about $1$-$2$ minutes during which the user is idle, and so smaller batch sizes lead to longer time investment from the users. As a good compromise, we chose $100$ as our batch size.

\vspace{1ex}
\noindent\textbf{Stronger pretrained model improves performance.}
Since our system leverages image-text co-embeddings to find relevant images and quickly train classifiers, a logical question is: how does changing the underlying embedding change the performance of the classifier? To do this, we compare CLIP versus ALIGN as the underlying embedding by replacing our pre-cached CLIP embeddings with ALIGN. We find that, with ALIGN, the AUROC of the final Agile model increased from 0.72 to 0.80 with a relative gain of 11.5\%. The AUPR increased from 0.68 to 0.76, a relative gain of 13.1\%. Furthermore, as Figure~\ref{fig:results_per_concept-auc-pr} demonstrates, both the ALIGN zero-shot and Agile models outperform their CLIP counterparts for almost every concept. This shows that building stronger image-text co-embeddings is foundational to improving the Agile Modeling process.

\section{Experiments with ImageNet21k}
Our user study validates the Agile Modeling framework on a small number of concepts over a web-scale unlabelled dataset. Now, we confirm that our framework can be effectively applied across a larger number of concepts to achieve significant improvements over zero-shot baselines. Due to the scale of this experiment, we simulate the user annotations using a fully-labeled dataset.

\vspace{1ex}
\noindent\textbf{Experimental setup.}
We use the ImageNet21k dataset~\cite{deng_imagenet_2009} which contains $21$k classes and over $14$M images. Out of these we select a subset of both easy and difficult classes, as described below. Each class corresponds to a binary classification problem as before. We apply the Agile Modeling framework with the ImageNet21k training set as the unlabeled data pool, and the test set for evaluation. Ground-truth class labels included in the dataset simulate a user providing ratings. Since the Agile Modeling process starts at concept definition with no labeled data, we use the class name and its corresponding WordNet~\cite{miller1995wordnet} description as positive text phrases in the text-to-image expansion step.
As before, we use a batch size of $100$ and $5$ rounds of active learning. We use ALIGN embeddings.

\vspace{1ex}
\noindent\textbf{Concept selection.}
We use a subset of $100$ of the $21$k concepts for evaluation. $50$ ``easy" concepts are selected at random from the ImageNet $1000$ class list. Additionally, we aim to replicate the ambiguity and difficulty of our original concepts by carefully selecting 50 further concepts with the following criteria based the WordNet lexicographical hierarchy: (1) $2$-$20$ hyponyms, to ensure visual variety, (2) more than $1$ lemma, to ensure ambiguity, (3) not an animal or plant, which have objective descriptions. Of the $546$ remaining concepts, our $50$ ``hard" concepts are selected at random. The full list of chosen concepts is in Appendix~\ref{appendix:imagenet}.

\begin{figure}[t]
  \centering
  \vspace{-2ex}
   \includegraphics[width=0.9\linewidth]{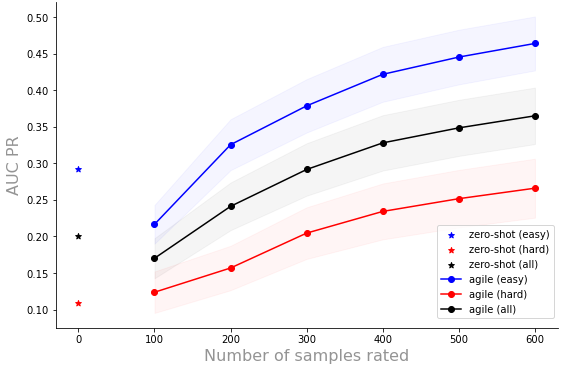}
   \vspace{-1em}
   \caption{Model performance per amount of samples on ImageNet21k for both easy and hard classes (AUC PR mean and std error over classes). Each $\bullet$ represents an AL round.}
   \label{fig:in21k-auc-pr}
  \vspace{-3ex}
\end{figure}

\vspace{1ex}
\noindent\textbf{Results.}
In Figure~\ref{fig:in21k-auc-pr} we show the results of applying the Agile Modeling framework to ImageNet21k. We see a similar trend to our user experiments, with significant improvements over zero-shot baselines as well as continued improvement with each active learning round. We further observe that the ``easy" concepts attained higher scores after the Agile Modeling process than the ``hard" concepts. The zero-shot baseline differed significantly between the ``easy" and ``hard" concepts with scores of 0.29 and 0.11, respectively. The equivalent of 30 minutes of human work yields a 20\% boost in AUC PR over the zero-shot baseline.
\section{Discussion \& conclusion}
\label{sec:discussion}

We formalized the Agile Modeling problem to turn any visual concept from an idea into a trained image classifier. We promote the notion of incorporating the user-in-the-loop, by supporting users with interactions that do not require any machine learning experience. We show that by using the latest advances in image-text pretrained models, we are able to initialize, train, and perform active learning in just a few minutes, enabling real-time user interaction for rapid model creation in less than 30 minutes.
Via a simple prototype, we demonstrate the value of users over crowd labelers in generating classifiers for subjective user-defined concepts.
We hope that our work showcases the opportunities and challenges of Agile Modeling and encourages future efforts.

{\small
\bibliographystyle{ieee_fullname}
\bibliography{ms}

\begin{thebibliography}{10}\itemsep=-1pt

\bibitem{amershi2014power}
Saleema Amershi, Maya Cakmak, William~Bradley Knox, and Todd Kulesza.
\newblock Power to the people: The role of humans in interactive machine
  learning.
\newblock {\em Ai Magazine}, 35(4):105--120, 2014.

\bibitem{ash2019deep}
Jordan~T Ash, Chicheng Zhang, Akshay Krishnamurthy, John Langford, and Alekh
  Agarwal.
\newblock Deep batch active learning by diverse, uncertain gradient lower
  bounds.
\newblock In {\em International Conference on Learning Representations}, 2019.

\bibitem{aumasson2012siphash}
Jean-Philippe Aumasson and Daniel~J Bernstein.
\newblock Siphash: a fast short-input prf.
\newblock In {\em International Conference on Cryptology in India}, pages
  489--508. Springer, 2012.

\bibitem{chen2020big}
Ting Chen, Simon Kornblith, Kevin Swersky, Mohammad Norouzi, and Geoffrey~E
  Hinton.
\newblock Big self-supervised models are strong semi-supervised learners.
\newblock {\em Advances in neural information processing systems},
  33:22243--22255, 2020.

\bibitem{Chuang}
Galen Chuang, Giulia DeSalvo, Lazaros Karydas, Jean-Francois Kagy, Afshin
  Rostamizadeh, and A Theeraphol.
\newblock Active learning empirical study.
\newblock In {\em NeurIPS 2019 Workshop on Learning with Rich Experience:
  Integration of Learning Paradigms}, 2019.

\bibitem{citovsky2021}
Gui Citovsky, Giulia DeSalvo, Claudio Gentile, Lazaros Karydas, Anand
  Rajagopalan, Afshin Rostamizadeh, and Sanjiv Kumar.
\newblock Batch active learning at scale.
\newblock In {\em Advances in Neural Information Processing Systems}, 2021.

\bibitem{cohen2022my}
Niv Cohen, Rinon Gal, Eli~A Meirom, Gal Chechik, and Yuval Atzmon.
\newblock “this is my unicorn, fluffy”: Personalizing frozen
  vision-language representations.
\newblock In {\em Computer Vision--ECCV 2022: 17th European Conference, Tel
  Aviv, Israel, October 23--27, 2022, Proceedings, Part XX}, pages 558--577.
  Springer, 2022.

\bibitem{coleman2022similarity}
Cody Coleman, Edward Chou, Julian Katz-Samuels, Sean Culatana, Peter Bailis,
  Alexander~C Berg, Robert Nowak, Roshan Sumbaly, Matei Zaharia, and I~Zeki
  Yalniz.
\newblock Similarity search for efficient active learning and search of rare
  concepts.
\newblock In {\em Proceedings of the AAAI Conference on Artificial
  Intelligence}, volume~36, pages 6402--6410, 2022.

\bibitem{coleman2019selection}
Cody Coleman, Christopher Yeh, Stephen Mussmann, Baharan Mirzasoleiman, Peter
  Bailis, Percy Liang, Jure Leskovec, and Matei Zaharia.
\newblock Selection via proxy: Efficient data selection for deep learning.
\newblock {\em arXiv preprint arXiv:1906.11829}, 2019.

\bibitem{culotta2005reducing}
Aron Culotta and Andrew McCallum.
\newblock Reducing labeling effort for structured prediction tasks.
\newblock In {\em AAAI}, volume~5, pages 746--751, 2005.

\bibitem{daum2022vocal}
Maureen Daum, Enhao Zhang, Dong He, Magdalena Balazinska, Brandon Haynes,
  Ranjay Krishna, Apryle Craig, and Aaron Wirsing.
\newblock Vocal: Video organization and interactive compositional analytics.
\newblock In {\em 12th Annual Conference on Innovative Data Systems Research
  (CIDR’22)}, 2022.

\bibitem{daum2023vocalexplore}
Maureen Daum, Enhao Zhang, Dong He, Brandon Haynes, Ranjay Krishna, and
  Magdalena Balazinska.
\newblock Vocalexplore: Pay-as-you-go video data exploration and model
  building.
\newblock {\em arXiv preprint arXiv:2301.00929}, 2023.

\bibitem{deng_imagenet_2009}
Jia Deng, Wei Dong, Richard Socher, Li-Jia Li, Kai Li, and Li Fei-Fei.
\newblock Imagenet: A large-scale hierarchical image database.
\newblock In {\em Proceedings of the IEEE/CVF Conference on Computer Vision and
  Pattern Recognition}, 2009.

\bibitem{dow2012shepherding}
Steven Dow, Anand Kulkarni, Scott Klemmer, and Bj{\"o}rn Hartmann.
\newblock Shepherding the crowd yields better work.
\newblock In {\em Proceedings of the ACM 2012 conference on computer supported
  cooperative work}, pages 1013--1022, 2012.

\bibitem{fails2003interactive}
Jerry~Alan Fails and Dan~R Olsen~Jr.
\newblock Interactive machine learning.
\newblock In {\em Proceedings of the 8th international conference on
  Intelligent user interfaces}, pages 39--45, 2003.

\bibitem{fei2022searching}
Li Fei-Fei and Ranjay Krishna.
\newblock Searching for computer vision north stars.
\newblock {\em Daedalus}, 151(2):85--99, 2022.

\bibitem{fiebrink2009play}
Rebecca Fiebrink, Perry~R Cook, and Dan Trueman.
\newblock Play-along mapping of musical controllers.
\newblock In {\em ICMC}, 2009.

\bibitem{finn2017model}
Chelsea Finn, Pieter Abbeel, and Sergey Levine.
\newblock Model-agnostic meta-learning for fast adaptation of deep networks.
\newblock In {\em International Conference on Machine Learning}, pages
  1126--1135. PMLR, 2017.

\bibitem{gadiraju2017clarity}
Ujwal Gadiraju, Jie Yang, and Alessandro Bozzon.
\newblock Clarity is a worthwhile quality: On the role of task clarity in
  microtask crowdsourcing.
\newblock In {\em Proceedings of the 28th ACM conference on hypertext and
  social media}, pages 5--14, 2017.

\bibitem{gal2022image}
Rinon Gal, Yuval Alaluf, Yuval Atzmon, Or Patashnik, Amit~H Bermano, Gal
  Chechik, and Daniel Cohen-Or.
\newblock An image is worth one word: Personalizing text-to-image generation
  using textual inversion.
\newblock {\em arXiv preprint arXiv:2208.01618}, 2022.

\bibitem{gordon2021disagreement}
Mitchell~L Gordon, Kaitlyn Zhou, Kayur Patel, Tatsunori Hashimoto, and
  Michael~S Bernstein.
\newblock The disagreement deconvolution: Bringing machine learning performance
  metrics in line with reality.
\newblock In {\em Proceedings of the 2021 CHI Conference on Human Factors in
  Computing Systems}, pages 1--14, 2021.

\bibitem{guo2016quantization}
Ruiqi Guo, Sanjiv Kumar, Krzysztof Choromanski, and David Simcha.
\newblock Quantization based fast inner product search.
\newblock In {\em Artificial intelligence and statistics}, pages 482--490.
  PMLR, 2016.

\bibitem{hochreiter2001learning}
Sepp Hochreiter, A~Steven Younger, and Peter~R Conwell.
\newblock Learning to learn using gradient descent.
\newblock In {\em International Conference on Artificial Neural Networks},
  pages 87--94. Springer, 2001.

\bibitem{jia2021scaling}
Chao Jia, Yinfei Yang, Ye Xia, Yi-Ting Chen, Zarana Parekh, Hieu Pham, Quoc Le,
  Yun-Hsuan Sung, Zhen Li, and Tom Duerig.
\newblock Scaling up visual and vision-language representation learning with
  noisy text supervision.
\newblock In {\em International Conference on Machine Learning}, pages
  4904--4916. PMLR, 2021.

\bibitem{karamcheti2021mind}
Siddharth Karamcheti, Ranjay Krishna, Li Fei-Fei, and Christopher~D Manning.
\newblock Mind your outliers! investigating the negative impact of outliers on
  active learning for visual question answering.
\newblock In {\em Proceedings of the 59th Annual Meeting of the Association for
  Computational Linguistics and the 11th International Joint Conference on
  Natural Language Processing (Volume 1: Long Papers)}, pages 7265--7281, 2021.

\bibitem{khan2021personalizing}
Mina Khan, P Srivatsa, Advait Rane, Shriram Chenniappa, Asadali Hazariwala, and
  Pattie Maes.
\newblock Personalizing pre-trained models.
\newblock {\em arXiv preprint arXiv:2106.01499}, 2021.

\bibitem{adam}
Diederik~P. Kingma and Jimmy Ba.
\newblock Adam: {A} method for stochastic optimization.
\newblock In Yoshua Bengio and Yann LeCun, editors, {\em 3rd International
  Conference on Learning Representations, {ICLR} 2015, San Diego, CA, USA, May
  7-9, 2015, Conference Track Proceedings}, 2015.

\bibitem{knox2013learning}
W~Bradley Knox and Peter Stone.
\newblock Learning non-myopically from human-generated reward.
\newblock In {\em Proceedings of the 2013 international conference on
  Intelligent user interfaces}, pages 191--202, 2013.

\bibitem{openimages}
Ivan Krasin, Tom Duerig, Neil Alldrin, Andreas Veit, Sami Abu-El-Haija, Serge
  Belongie, David Cai, Zheyun Feng, Vittorio Ferrari, Victor Gomes, Abhinav
  Gupta, Dhyanesh Narayanan, Chen Sun, Gal Chechik, and Kevin Murphy.
\newblock Openimages: A public dataset for large-scale multi-label and
  multi-class image classification.
\newblock {\em Dataset available from https://github.com/openimages}, 2016.

\bibitem{krishna2021visual}
Ranjay Krishna, Mitchell Gordon, Li Fei-Fei, and Michael Bernstein.
\newblock Visual intelligence through human interaction.
\newblock {\em Artificial Intelligence for Human Computer Interaction: A Modern
  Approach}, pages 257--314, 2021.

\bibitem{krishna2022socially}
Ranjay Krishna, Donsuk Lee, Li Fei-Fei, and Michael~S Bernstein.
\newblock Socially situated artificial intelligence enables learning from human
  interaction.
\newblock {\em Proceedings of the National Academy of Sciences},
  119(39):e2115730119, 2022.

\bibitem{lease2011quality}
Matthew Lease.
\newblock On quality control and machine learning in crowdsourcing.
\newblock In {\em Workshops at the Twenty-Fifth AAAI Conference on Artificial
  Intelligence}. Citeseer, 2011.

\bibitem{uncertainty}
David Lewis and William Gale.
\newblock A sequential algorithm for training text classifiers.
\newblock In {\em ACM SIGIR Conference on Research and Development in
  Information Retrieval}, 1994.

\bibitem{lewis1995sequential}
David~D Lewis.
\newblock A sequential algorithm for training text classifiers: Corrigendum and
  additional data.
\newblock In {\em Acm Sigir Forum}, volume~29, pages 13--19. ACM New York, NY,
  USA, 1995.

\bibitem{lin2014microsoft}
Tsung-Yi Lin, Michael Maire, Serge Belongie, James Hays, Pietro Perona, Deva
  Ramanan, Piotr Doll{\'a}r, and C~Lawrence Zitnick.
\newblock Microsoft {COCO:} common objects in context.
\newblock In {\em European Conference on Computer Vision}, pages 740--755.
  Springer, 2014.

\bibitem{loftin2016learning}
Robert Loftin, Bei Peng, James MacGlashan, Michael~L Littman, Matthew~E Taylor,
  Jeff Huang, and David~L Roberts.
\newblock Learning behaviors via human-delivered discrete feedback: modeling
  implicit feedback strategies to speed up learning.
\newblock {\em Autonomous agents and multi-agent systems}, 30:30--59, 2016.

\bibitem{miller1995wordnet}
George~A Miller.
\newblock Wordnet: a lexical database for english.
\newblock {\em Communications of the ACM}, 38(11):39--41, 1995.

\bibitem{mullapudi2021background}
Ravi~Teja Mullapudi, Fait Poms, William~R Mark, Deva Ramanan, and Kayvon
  Fatahalian.
\newblock Background splitting: Finding rare classes in a sea of background.
\newblock In {\em Proceedings of the IEEE/CVF Conference on Computer Vision and
  Pattern Recognition}, pages 8043--8052, 2021.

\bibitem{Mullapudi}
Ravi~Teja Mullapudi, Fait Poms, William~R Mark, Deva Ramanan, and Kayvon
  Fatahalian.
\newblock Learning rare category classifiers on a tight labeling budget.
\newblock In {\em IEEE/CVF International Conference on Computer Vision}, pages
  8423--8432, 2021.

\bibitem{ouyang2022training}
Long Ouyang, Jeff Wu, Xu Jiang, Diogo Almeida, Carroll~L Wainwright, Pamela
  Mishkin, Chong Zhang, Sandhini Agarwal, Katarina Slama, Alex Ray, et~al.
\newblock Training language models to follow instructions with human feedback.
\newblock {\em arXiv preprint arXiv:2203.02155}, 2022.

\bibitem{park2019ai}
Junwon Park, Ranjay Krishna, Pranav Khadpe, Li Fei-Fei, and Michael Bernstein.
\newblock Ai-based request augmentation to increase crowdsourcing
  participation.
\newblock In {\em Proceedings of the AAAI Conference on Human Computation and
  Crowdsourcing}, volume~7, pages 115--124, 2019.

\bibitem{patel2010gestalt}
Kayur Patel, Naomi Bancroft, Steven~M Drucker, James Fogarty, Amy~J Ko, and
  James Landay.
\newblock Gestalt: integrated support for implementation and analysis in
  machine learning.
\newblock In {\em Proceedings of the 23nd annual ACM symposium on User
  interface software and technology}, pages 37--46, 2010.

\bibitem{tropel}
Genevieve Patterson, Grant Van~Horn, Serge Belongie, Pietro Perona, and James
  Hays.
\newblock Tropel: Crowdsourcing detectors with minimal training.
\newblock In {\em Third AAAI Conference on Human Computation and
  Crowdsourcing}, 2015.

\bibitem{pinsler2019bayesian}
Robert Pinsler, Jonathan Gordon, Eric Nalisnick, and Jos{\'e}~Miguel
  Hern{\'a}ndez-Lobato.
\newblock Bayesian batch active learning as sparse subset approximation.
\newblock {\em Advances in neural information processing systems}, 32, 2019.

\bibitem{pratt2022does}
Sarah Pratt, Rosanne Liu, and Ali Farhadi.
\newblock What does a platypus look like? generating customized prompts for
  zero-shot image classification.
\newblock {\em arXiv preprint arXiv:2209.03320}, 2022.

\bibitem{radford2021learning}
Alec Radford, Jong~Wook Kim, Chris Hallacy, Aditya Ramesh, Gabriel Goh,
  Sandhini Agarwal, Girish Sastry, Amanda Askell, Pamela Mishkin, Jack Clark,
  et~al.
\newblock Learning transferable visual models from natural language
  supervision.
\newblock In {\em International Conference on Machine Learning}, pages
  8748--8763. PMLR, 2021.

\bibitem{Snorkel}
A Ratner, S.H Bach, H Ehrenberg, J Fries, S Wu, and C Re.
\newblock Snorkel: Rapid training data creation with weak supervision.
\newblock In {\em VLDB Endowment}, pages 269--282, 2017.

\bibitem{ren2021survey}
Pengzhen Ren, Yun Xiao, Xiaojun Chang, Po-Yao Huang, Zhihui Li, Brij~B Gupta,
  Xiaojiang Chen, and Xin Wang.
\newblock A survey of deep active learning.
\newblock {\em ACM computing surveys (CSUR)}, 54(9):1--40, 2021.

\bibitem{russakovsky2015imagenet}
Olga Russakovsky, Jia Deng, Hao Su, Jonathan Krause, Sanjeev Satheesh, Sean Ma,
  Zhiheng Huang, Andrej Karpathy, Aditya Khosla, Michael Bernstein, et~al.
\newblock Imagenet large scale visual recognition challenge.
\newblock {\em International journal of computer vision}, 115(3):211--252,
  2015.

\bibitem{margin_sampling}
Decomain~C Scheffer, T and S Wrobel.
\newblock Active hidden markov models for information extraction.
\newblock In {\em International Conference on Advances in Intelligent Data
  Analysis (CAIDA)}, page 309–318, 2001.

\bibitem{scheffer2001active}
Tobias Scheffer, Christian Decomain, and Stefan Wrobel.
\newblock Active hidden markov models for information extraction.
\newblock In {\em International Symposium on Intelligent Data Analysis}, pages
  309--318. Springer, 2001.

\bibitem{schuhmann2022laion}
Christoph Schuhmann, Romain Beaumont, Richard Vencu, Cade Gordon, Ross
  Wightman, Mehdi Cherti, Theo Coombes, Aarush Katta, Clayton Mullis, Mitchell
  Wortsman, et~al.
\newblock Laion-5b: An open large-scale dataset for training next generation
  image-text models.
\newblock {\em arXiv preprint arXiv:2210.08402}, 2022.

\bibitem{schuhmann2021laion}
Christoph Schuhmann, Richard Vencu, Romain Beaumont, Robert Kaczmarczyk,
  Clayton Mullis, Aarush Katta, Theo Coombes, Jenia Jitsev, and Aran
  Komatsuzaki.
\newblock Laion-400m: Open dataset of clip-filtered 400 million image-text
  pairs.
\newblock {\em arXiv preprint arXiv:2111.02114}, 2021.

\bibitem{sener2017active}
Ozan Sener and Silvio Savarese.
\newblock Active learning for convolutional neural networks: A core-set
  approach.
\newblock {\em arXiv preprint arXiv:1708.00489}, 2017.

\bibitem{al_survey}
Burr Settles.
\newblock Active learning literature survey.
\newblock {\em computer sciences technical report 1648, University of
  Wisconsin, Madison}, 2010.

\bibitem{shen2017deep}
Yanyao Shen, Hyokun Yun, Zachary~C Lipton, Yakov Kronrod, and Animashree
  Anandkumar.
\newblock Deep active learning for named entity recognition.
\newblock {\em arXiv preprint arXiv:1707.05928}, 2017.

\bibitem{sheng2008get}
Victor~S Sheng, Foster Provost, and Panagiotis~G Ipeirotis.
\newblock Get another label? improving data quality and data mining using
  multiple, noisy labelers.
\newblock In {\em Proceedings of the 14th ACM SIGKDD international conference
  on Knowledge discovery and data mining}, pages 614--622, 2008.

\bibitem{snell2017prototypical}
Jake Snell, Kevin Swersky, and Richard Zemel.
\newblock Prototypical networks for few-shot learning.
\newblock {\em Advances in Neural Information Processing Systems}, 30, 2017.

\bibitem{thomaz2008teachable}
Andrea~L Thomaz and Cynthia Breazeal.
\newblock Teachable robots: Understanding human teaching behavior to build more
  effective robot learners.
\newblock {\em Artificial Intelligence}, 172(6-7):716--737, 2008.

\bibitem{thomee2016yfcc100m}
Bart Thomee, David~A Shamma, Gerald Friedland, Benjamin Elizalde, Karl Ni,
  Douglas Poland, Damian Borth, and Li-Jia Li.
\newblock Yfcc100m: The new data in multimedia research.
\newblock {\em Communications of the ACM}, 59(2):64--73, 2016.

\bibitem{tian2020rethinking}
Yonglong Tian, Yue Wang, Dilip Krishnan, Joshua~B Tenenbaum, and Phillip Isola.
\newblock Rethinking few-shot image classification: a good embedding is all you
  need?
\newblock In {\em European Conference on Computer Vision}, pages 266--282.
  Springer, 2020.

\bibitem{metadataset}
Eleni Triantafillou, Tyler Zhu, Vincent Dumoulin, Pascal Lamblin, Utku Evci,
  Kelvin Xu, Ross Goroshin, Carles Gelada, Kevin Swersky, Pierre-Antoine
  Manzagol, et~al.
\newblock Meta-dataset: A dataset of datasets for learning to learn from few
  examples.
\newblock {\em arXiv preprint arXiv:1903.03096}, 2019.

\bibitem{varma2017inferring}
Paroma Varma, Bryan~D He, Payal Bajaj, Nishith Khandwala, Imon Banerjee, Daniel
  Rubin, and Christopher R{\'e}.
\newblock Inferring generative model structure with static analysis.
\newblock {\em Advances in neural information processing systems}, 30, 2017.

\bibitem{vinyals2016matching}
Oriol Vinyals, Charles Blundell, Timothy Lillicrap, Daan Wierstra, et~al.
\newblock Matching networks for one shot learning.
\newblock {\em Advances in neural information processing systems}, 29, 2016.

\bibitem{wang2020generalizing}
Yaqing Wang, Quanming Yao, James~T Kwok, and Lionel~M Ni.
\newblock Generalizing from a few examples: A survey on few-shot learning.
\newblock {\em ACM Computing Surveys (CSUR)}, 53(3):1--34, 2020.

\bibitem{wu2017multiscale}
Xiang Wu, Ruiqi Guo, Ananda~Theertha Suresh, Sanjiv Kumar, Daniel~N
  Holtmann-Rice, David Simcha, and Felix Yu.
\newblock Multiscale quantization for fast similarity search.
\newblock {\em Advances in neural information processing systems}, 30, 2017.

\bibitem{zhang2023equi}
Enhao Zhang, Maureen Daum, Dong He, Brandon Haynes, Ranjay Krishna, and
  Magdalena Balazinska.
\newblock Equi-vocal: Synthesizing queries for compositional video events from
  limited user interactions.
\newblock {\em arXiv preprint arXiv:2301.00929}, 2023.

\end{thebibliography}
}

\clearpage

\appendix

\section{Concepts}
\label{appendix:concepts}

We provide the full list of concepts, along with the text phrases provided by the users. Each concept name was automatically added to the list of positive text phrases.

\begin{enumerate}[noitemsep]
    \item \concept{gourmet tuna}
    \begin{enumerate}
        \item Positive text phrases: tuna sushi, seared tuna, tuna sashimi
        \item Negative text phrases: canned tuna, tuna sandwich, tuna fish, tuna fishing
    \end{enumerate}
    \item \concept{emergency service}
    \begin{enumerate}
        \item Positive text phrases: firefighting, paramedic, ambulance, disaster worker, search and rescue
        \item Negative text phrases: construction, crossing guard, military
    \end{enumerate}
    \item \concept{healthy dish}
    \begin{enumerate}
        \item Positive text phrases: salad, fish dish, vegetables, healthy food
        \item Negative text phrases: fast food, fried food, sugary food, fatty food
    \end{enumerate}
    \item \concept{in-ear headphones}
    \begin{enumerate}
        \item Positive text phrases: in-ear headphones, airpods, earbuds
        \item Negative text phrases: earrings, bone headphones, over-ear headphones
    \end{enumerate}
    \item \concept{hair coloring}
    \begin{enumerate}
        \item Positive text phrases: hair coloring service, hair coloring before and after
        \item Negative text phrases: hair coloring product
    \end{enumerate}
    \item \concept{arts and crafts}
    \begin{enumerate}
        \item Positive text phrases: kids crafts, scrapbooking, hand made decorations
        \item Negative text phrases: museum art, professional painting, sculptures
    \end{enumerate}
    \item \concept{home fragrance}
    \begin{enumerate}
        \item Positive text phrases: home fragrance flickr, scented candles, air freshener, air freshener flickr, room fragrance, room fragrance flickr, scent sachet, potpourri, potpourri flickr
        \item Negative text phrases: birthday candles, birthday candles flickr, religious candles, religious candles flickr, car freshener, car freshener flickr, perfume, perfume flickr
    \end{enumerate}
    \item \concept{single sneaker on white background}
    \begin{enumerate}
        \item Positive text phrases: one sneaker on white background
        \item Negative text phrases: two sneakers on white background, leather shoe
    \end{enumerate}
    \item \concept{dance}
    \begin{enumerate}
        \item Positive text phrases: ballet, tango, ballroom dancing, classical dancing, professional dance
        \item Negative text phrases: sports, fitness, zumba, ice skating
    \end{enumerate}
    \item \concept{hand pointing}
    \begin{enumerate}
        \item Positive text phrases: hand pointing, meeting with pointing hand, cartoon hand pointing, pointing at screen
        \item Negative text phrases: thumbs up, finger gesture, hands, sign language
    \end{enumerate}
    \item \concept{astronaut}
    \begin{enumerate}
        \item Positive text phrases: female astronaut, spacecraft crew, space traveler
        \item Negative text phrases: spacecraft, space warrior, scuba diver
    \end{enumerate}
    \item \concept{stop sign}
    \begin{enumerate}
        \item Positive text phrases: stop sign in traffic, stop sign held by a construction worker, stop sign on a bus, stop sign on the road, outdoor stop sign, stop sign in the wild 
        \item Negative text phrases: indoor stop sign, slow sign, traffic light sign, stop sign on a poster, stop sign on the wall, cartoon stop sign, stop sign only
    \end{enumerate}
    \item \concept{pie chart}
    \begin{enumerate}
        \item Positive text phrases: pie-chart
        \item Negative text phrases: pie, bar chart, plot
    \end{enumerate}
    \item \concept{block tower}
    \begin{enumerate}
        \item Positive text phrases: toy tower
        \item Negative text phrases: tower block, building
    \end{enumerate}
\end{enumerate}

\section{Crowd task design}
\label{appendix:crowd-task-design}

\begin{figure}
    \centering
    \includegraphics[width=\linewidth]{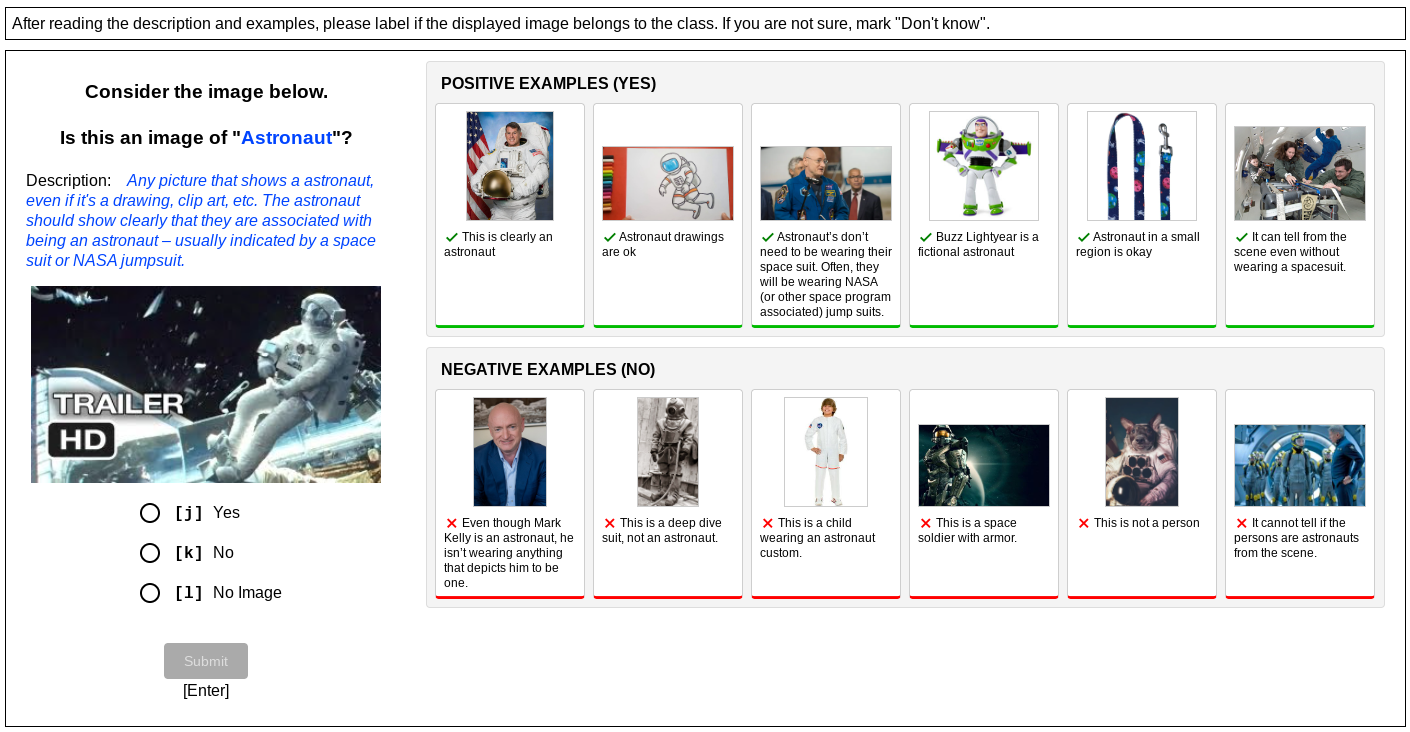}
    \caption{An example template we use for crowd labeling.}
    \label{fig:crowd-template}
\end{figure}

Crowd workers are onboarded to the binary image classification task then given batches of images to label, where each batch contains images from the same concept type to minimize cross-concept mislabeling. In Figure~\ref{fig:crowd-template} we show the task we present to crowd workers for image classification. The template contains the image to classify, as well as a description of the image concept and a set of positive and negative examples created by the user who created the concept. Each image is sent to three crowd workers and the label is decided by majority vote.


\begin{figure*}[!h]
  \centering
   \includegraphics[width=\textwidth]{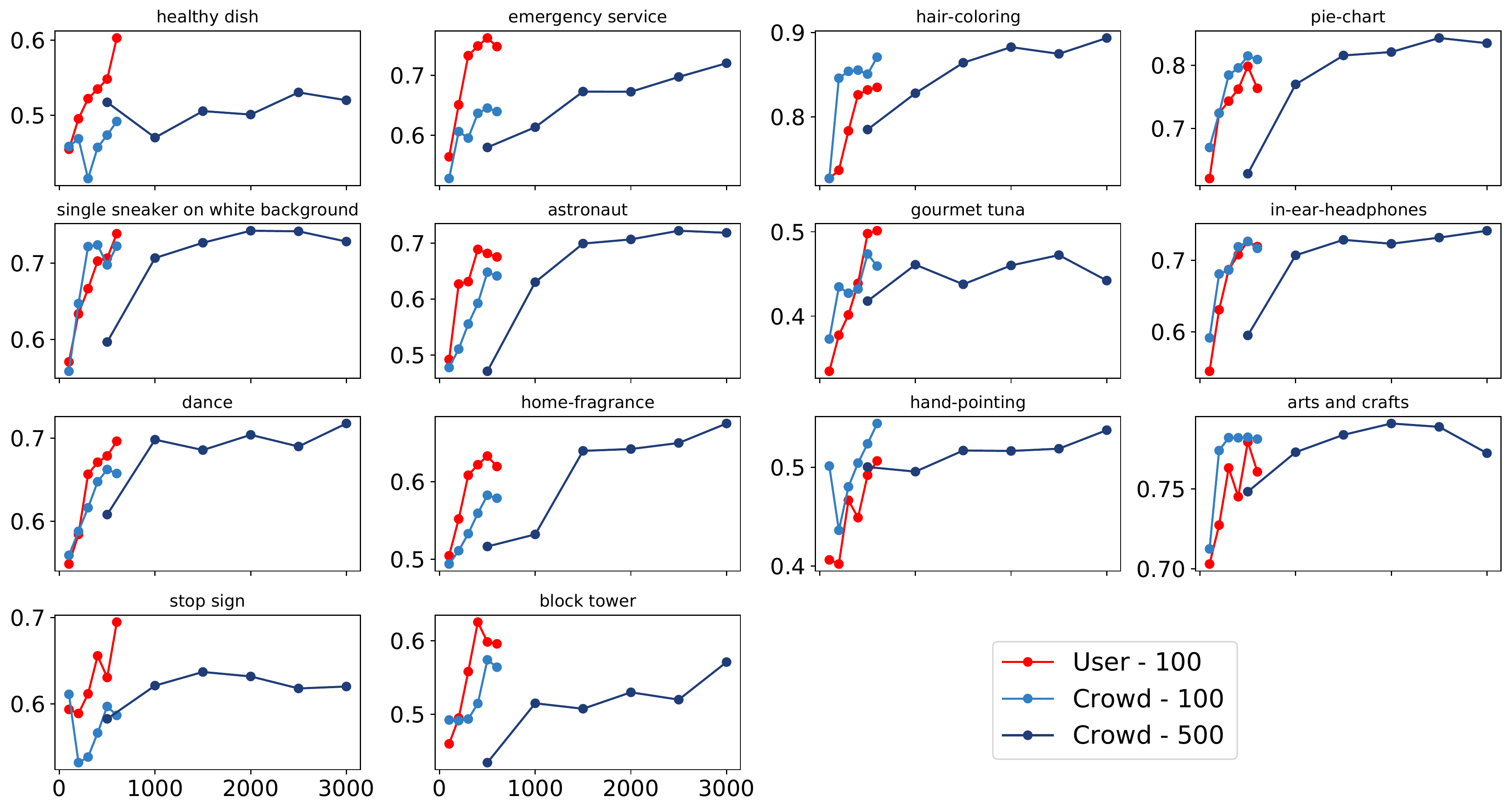}
   \vspace{-1em}
   \caption{Results per concept comparing user model performance versus crowd. We show the AUC PR (y-axis) per number of samples rated (x-axis) for each of the three active learning experimental settings:  user (batch size = 100), crowd (batch size = 100), and crowd (batch size = 500).}
   \label{fig:al-per-concept-expert-vs-crowd}
\end{figure*}

\section{Experimental details}
\label{appendix:experimental-details}

All models are trained using binary cross-entropy loss, a dropout rate of 0.5 and weight decay regularization with weight \num{1e-4}. We use the Adam optimizer \cite{adam} with learning rate \num{1e-4} and train for 10 epochs. To prevent overtriggering by the trained classifier, we sample 500k random images from the unlabeled set and automatically label them negative. During training, we upsample our labeled positives to be half the training set, while labeled negatives and the random negatives are each a quarter of the training set. All hyperparameters have been chosen on 2 held-out concepts.


\section{Active Learning}

\label{experiments:stanford}

\begin{figure}[!t]
  \centering
  \vspace{-2ex}
   \includegraphics[width=0.9\linewidth]{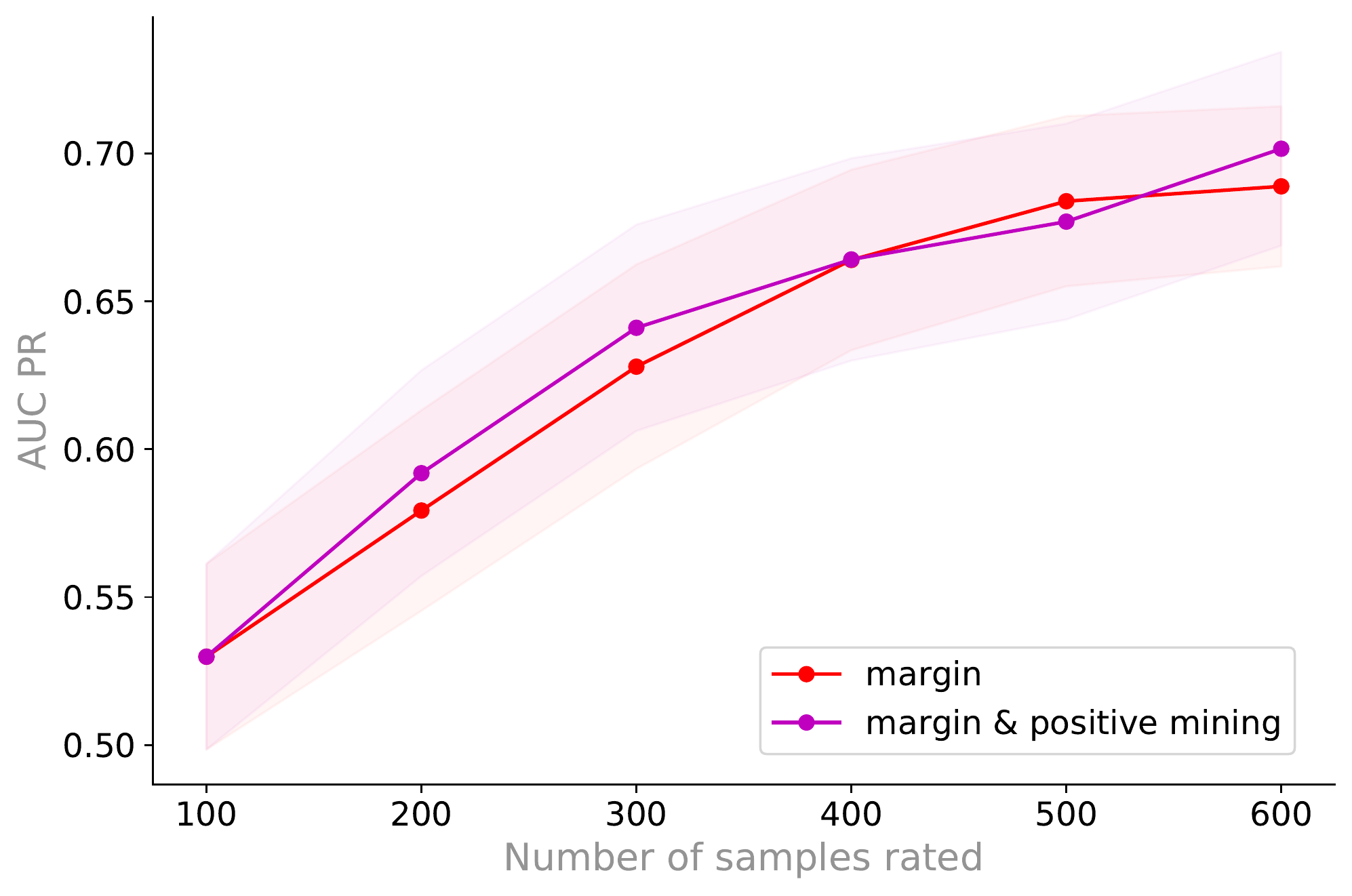}
    \vspace{-1ex}
   \caption{Model performance for two active learning methods: margin and the approach of \cite{Mullapudi} (margin \& positive mining). Each $\bullet$ corresponds to an AL round. We show the AUC PR mean and standard error over all concepts.}
   \label{fig:al_stanford}
   \vspace{-3ex}
\end{figure}

\textbf{Active learning method.}
Throughout the paper, we instantiate the active learning component with the well-known margin method~\cite{margin_sampling}. We now compare it to the active learning method used in Mullapudi et al \cite{Mullapudi}.  We ran a version of our instantiation of the Agile framework where we replace margin with the margin+positive mining strategy chosen by \cite{Mullapudi} and described in Section~\ref{sec:al}. The performance of the two methods per AL round is shown in Figure~\ref{fig:al_stanford}. Interestingly, despite the fact that Mullapudi et al. \cite{Mullapudi} introduced this hybrid approach to improve upon margin sampling, in this setting, on average, the two methods perform similarly across all AL rounds. We see the same effect on most concepts, when we inspect this on a per-concept basis in Appendix~\ref{appendix:al-results}.
One potential explanation for this is that the initial model trained before AL is already good enough (perhaps due to the powerful CLIP embeddings) for margin sampling to produce a dataset balanced in terms of positive and negative, and thus explicitly mining easy positives as in~\cite{Mullapudi} is not particularly useful.
Since the two methods perform equivalently, while margin being simpler and more efficient, we opted for margin in the rest of the experiments.

\section{Evaluation strategy}
\label{appendix:evaluation_strategy}
Because we are eliciting the concept from users, only they can correctly label every image. Therefore, when generating an evaluation set, the annotations must come from the user. However, since our users are real people with real time restrictions, this means that we cannot ask them to exhaustively rate a large evaluation set. We target less than 1000 images for each concept's evaluation set.

\subsection{Proposed evaluation strategies}
We considered the following strategies for evaluation:

\paragraph{Labeling the entire unlabeled set.} The most accurate evaluation metric is to label the entire unlabeled set. However, this is infeasible, as the user would have to label hundreds of millions of images.
\paragraph{Random sampling from unlabeled set.} To reduce the number of images to label, we could randomly sample until we hit a desired amount. However, since most of the concepts are rare ($<0.1\%$ of the total amount of data), this means our evaluation set would have very few positives.
\paragraph{Holdout of training data.} As the user labels new ground truth, hold out a fraction of it for evaluation. The benefit is that the user does not have to label any extra data. The main detriment is that the evaluation set comes from the exact same distribution as the training set, leading to overestimates of performance, as there are no new visual modes in the evaluation set.
\paragraph{Random sampling at fixed prediction frequencies.} Choose a set of operating points. For each operating point randomly sample K images with score higher than that operating point. The operating points can be selected as the model prediction frequency---for example, we can calculate precision of the highest confidence 100, 1000, and 10000 predictions. The metric that will be directly comparable across models is precision vs prediction frequency. To minimize rating cost we can use the deterministic hash approach. The main problem is that the choice of operating points varies depending on the particular class. Classes that are rare or harder to correctly predict may need stricter operating points than common and easy classes. Furthermore, with this approach we cannot compute a PR curve, just some metrics at specific operating points.
\paragraph{Stratified sampling without weights [our chosen approach].} Collect new evaluation images by (1) calculating model scores, (2) bucketing the images by model score (\eg, [0, 0.1), [0.1, 0.2),..., [0.8, 0.9), [0.9, 1]), (3) rating k examples per bucket. To minimize any bias towards any particular model, we can repeat this process to retrieve an evaluation set per model and merge to get the final evaluation set. Additionally, we can use a deterministic hash instead of random sampling to encourage high overlap across the images chosen to save on the total rating budget. The major upside is that, using a small number of images rated, we can get a relatively balanced dataset of positives and negatives, while also mining for hard examples to stress test the models. The main limitations of this method are:
\begin{enumerate}
    \item Stratified sampling requires good bucket boundaries to work well, which is not guaranteed.
    \item The metric will be biased since samples selected from buckets with a smaller number of candidates (such as the [0.9, 1] bucket) will have more influence than samples from buckets with lots of candidates (\eg the [0, 0.1) bucket).
    \item Merging image sets from multiple models may bias towards the models make common predictions. However, we hope that pseudorandom hashing selects the same images and prevents this from occurring.
\end{enumerate}

\paragraph{Stratified sampling with weights.} This involves the same process as stratified sampling without weights, but whenever computing a metric, you weigh the sample by the distribution of scores it came from. This unbiases sampling from each strata, but for very large buckets (\eg, the [0, 0.1) bucket), the weight would be extremely large. This means that predicting incorrectly on any of these images overpowers all correct predictions on other buckets.

Based on the pros and cons of all these approaches, we chose 
{\em stratified sampling without weights} for our experiments, which we believe is most representative for our problem setting.

\subsection{Evaluation set statistics}
In Table~\ref{tab:eval_dataset_stats}, we show that our stratified sampling method chooses a tractable number of images to rate, while keeping the positive and negative count relatively balanced.
\begin{table}[h!]
    \centering
    \begin{tabular}{l@{\hspace{1.2\tabcolsep}}c@{\hspace{1.2\tabcolsep}} c}
    \toprule
    \textbf{Concept Name} & \textbf{\# Images} & \textbf{Pos. Rate} \\
    \midrule
arts and crafts & 707 & 0.66 \\
astronaut & 637 & 0.36 \\
block tower & 669 & 0.36 \\
dance & 730 & 0.47 \\
emergency service & 675 & 0.50 \\
gourmet tuna & 576 & 0.27 \\
hair-coloring & 645 & 0.67 \\
hand-pointing & 832 & 0.34 \\
healthy dish & 633 & 0.36 \\
home-fragrance & 716 & 0.39 \\
in-ear-headphones & 687 & 0.42 \\
pie-chart & 594 & 0.42 \\
single sneaker on white background & 556 & 0.49 \\
stop sign & 704 & 0.44 \\
    \bottomrule
    \end{tabular}
    \caption{Statistics showing the number of images and the positive rate in each concept's evaluation set.}
    \label{tab:eval_dataset_stats}
    \vspace{-1em}
\end{table}

\section{User-in-the-loop vs crowd raters}
\label{appendix:crowd}

We include additional results comparing active learning with the user in the loop with active learning using crowd raters. 
Figure~\ref{fig:al-per-concept-expert-vs-crowd} shows detailed results, per concept, for the three experimental settings \texttt{User-100}, \texttt{Crowd-100} and \texttt{Crowd-500} described in Section~\ref{sec:value-of-domain-experts}.
We can notice how for difficult concepts (according to the difficulty scores in Appendix~\ref{appendix:concept-difficulty}) such as \concept{healthy dish}, the performance of the user models far exceeds that of the crowd raters, with far less samples. On the other hand, for easy concepts such as \concept{hair coloring} the models trained with more data from crowd raters end up superseding the best user model.

\begin{figure*}[!t]
     \centering
     \vspace{-2ex}
     \begin{subfigure}[b]{0.33\linewidth}
         \centering
         \includegraphics[width=\linewidth]{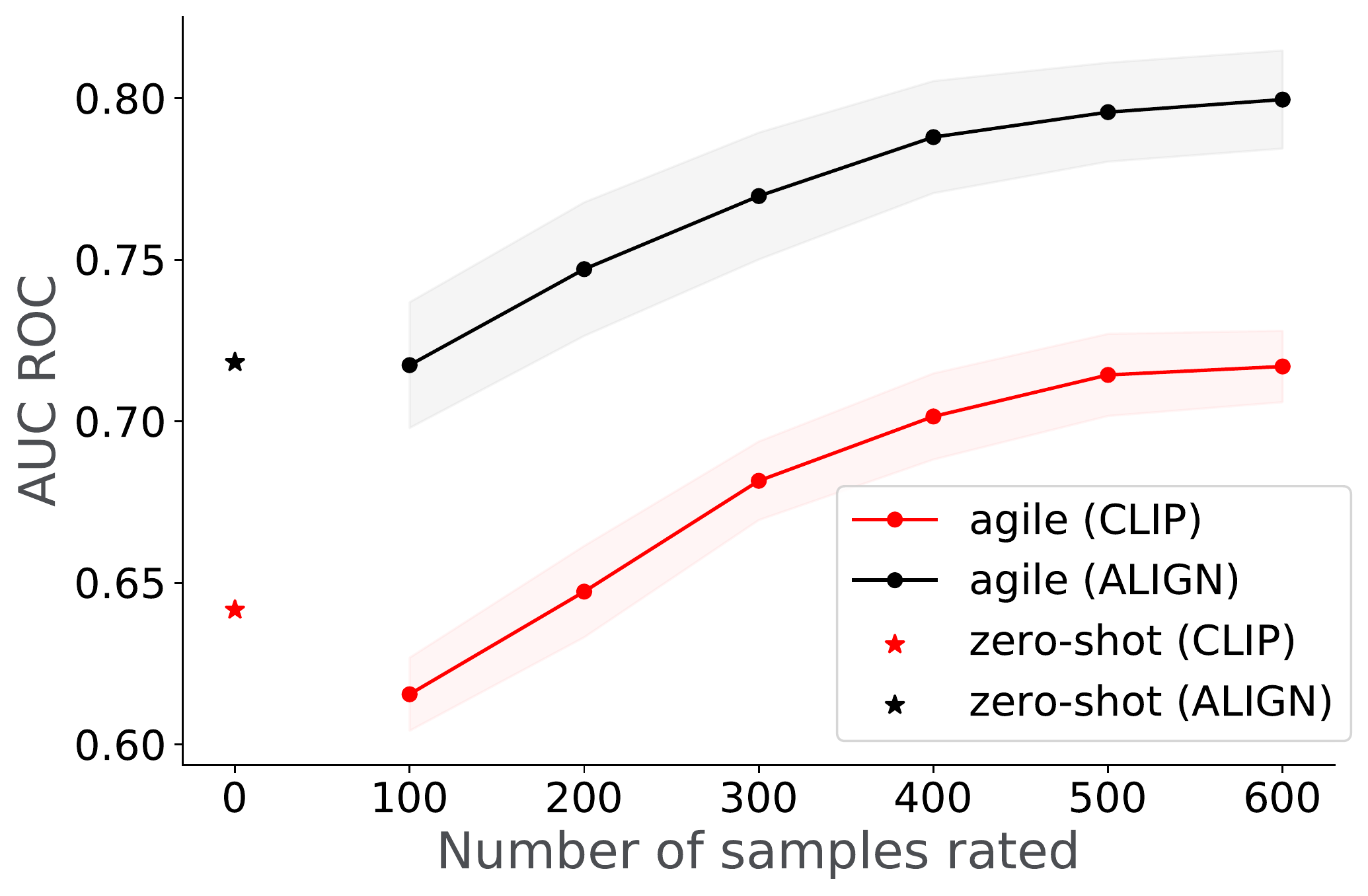}
         \caption{Area under the receiver-operator curve.}
         \label{fig:al-auc-roc}
     \end{subfigure}
     \hfill
     \begin{subfigure}[b]{0.33\linewidth}
         \centering
         \includegraphics[width=\linewidth]{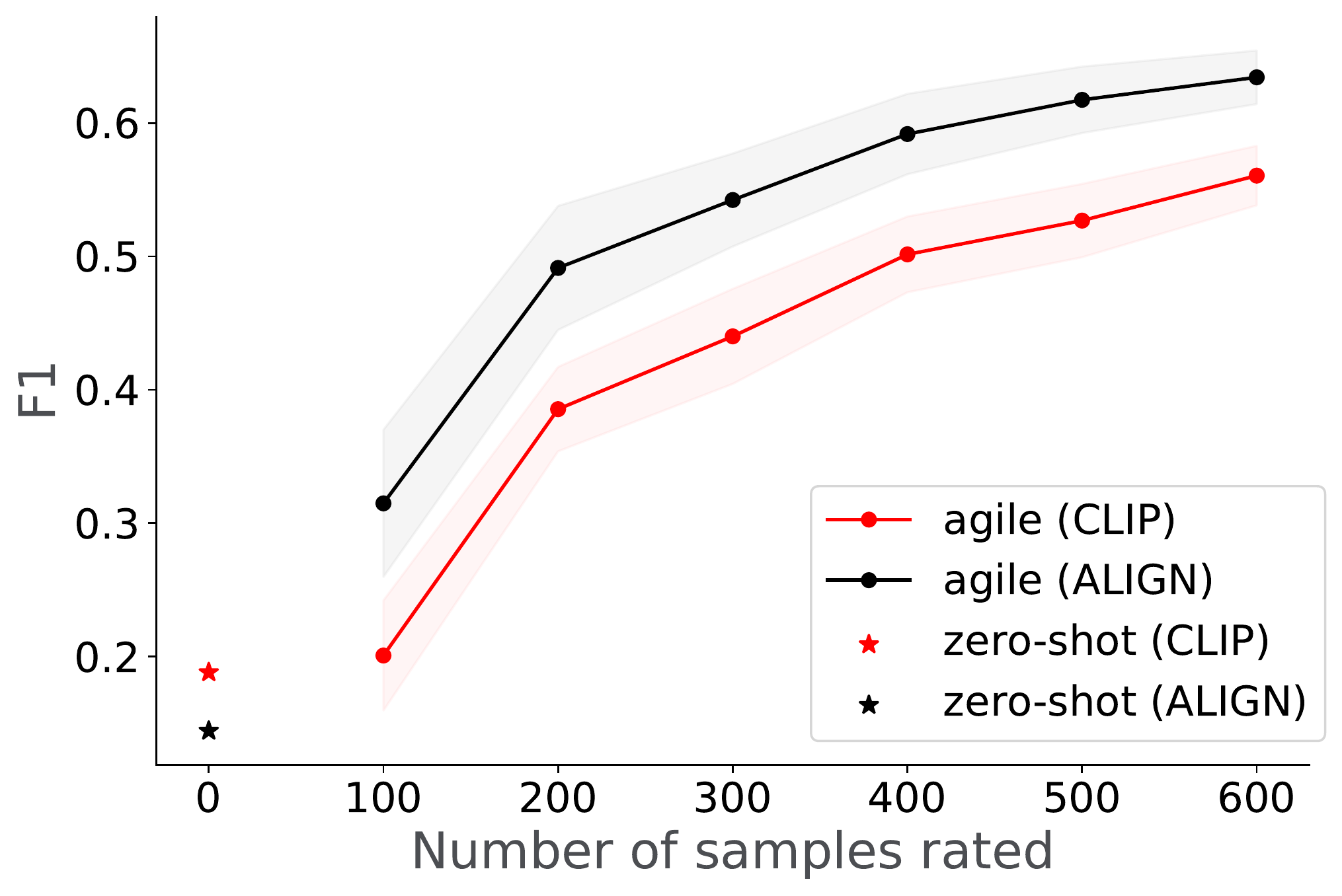}
         \caption{F1 score.}
         \label{fig:al-f1}
     \end{subfigure}
     \hfill
     \begin{subfigure}[b]{0.33\linewidth}
         \centering
         \includegraphics[width=\linewidth]{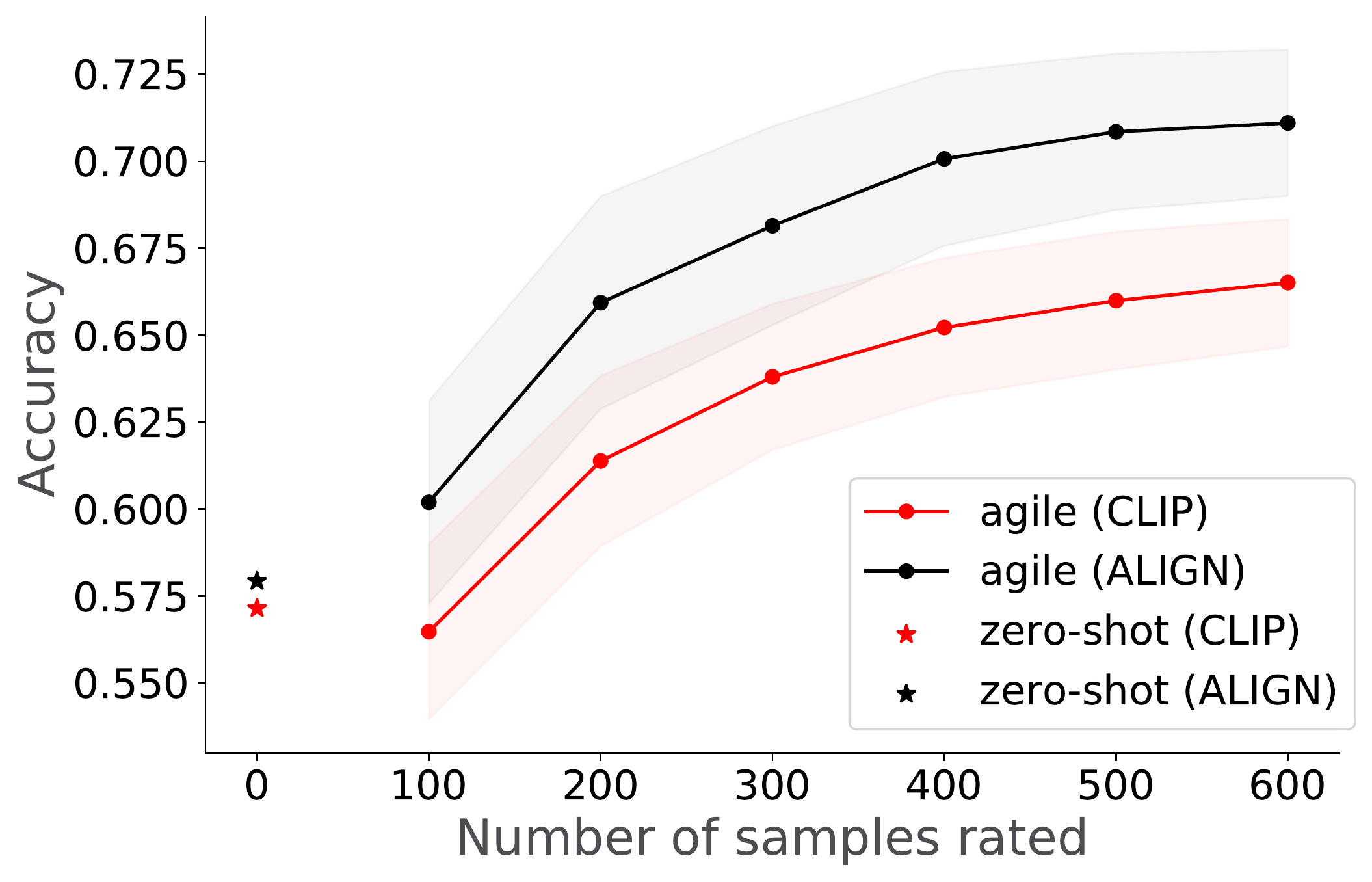}
         \caption{Accuracy.}
         \label{fig:al-acc}
     \end{subfigure}
     \caption{Model performance per amount of samples rated by the user. Mean and standard error over all concepts, for multiple metrics.}
    \label{fig:al-additional-metrics}
\end{figure*}

\section{Additional active learning results}
\label{appendix:al-results}

\subsection{Additional metrics}
We include here additional active learning results, measuring the amount of rating by user versus model performance. Figure~\ref{fig:al-additional-metrics} shows the results in terms of AUC ROC, F1 score, and accuracy.
Note that, unlike AUC PR and AUC ROC, for computing the F1 score and accuracy one must choose a threshold on the model prediction score that determines whether a sample is on the positive or negative side of the decision boundary. For our trained MLP models, we used the common 0.5 threshold. For the zero-shot models, the threshold 0.5 is not a good choice, because the cosine similarities for both positive and negative are often smaller than this. In fact, \cite{schuhmann2022laion} did an analysis of the right choice of threshold based on a human inspection on LAION-5B, and they recommend using the threshold 0.28 when using CLIP embeddings; we also use this threshold. 
We similarly chose 0.2 as a threshold when using ALIGN based on our own inspection.

Based on the results in Figure~\ref{fig:al-additional-metrics}, we noticed the same consistent observations with all metrics: (1) the performance increases with every active learning round; (2) the performance increase is  faster in the beginning, and starting to plateau in the later AL rounds; (3) the models based in ALIGN embeddings are consistently better than those using CLIP.



\begin{figure*}[!t]
  \centering
   \includegraphics[width=\textwidth]{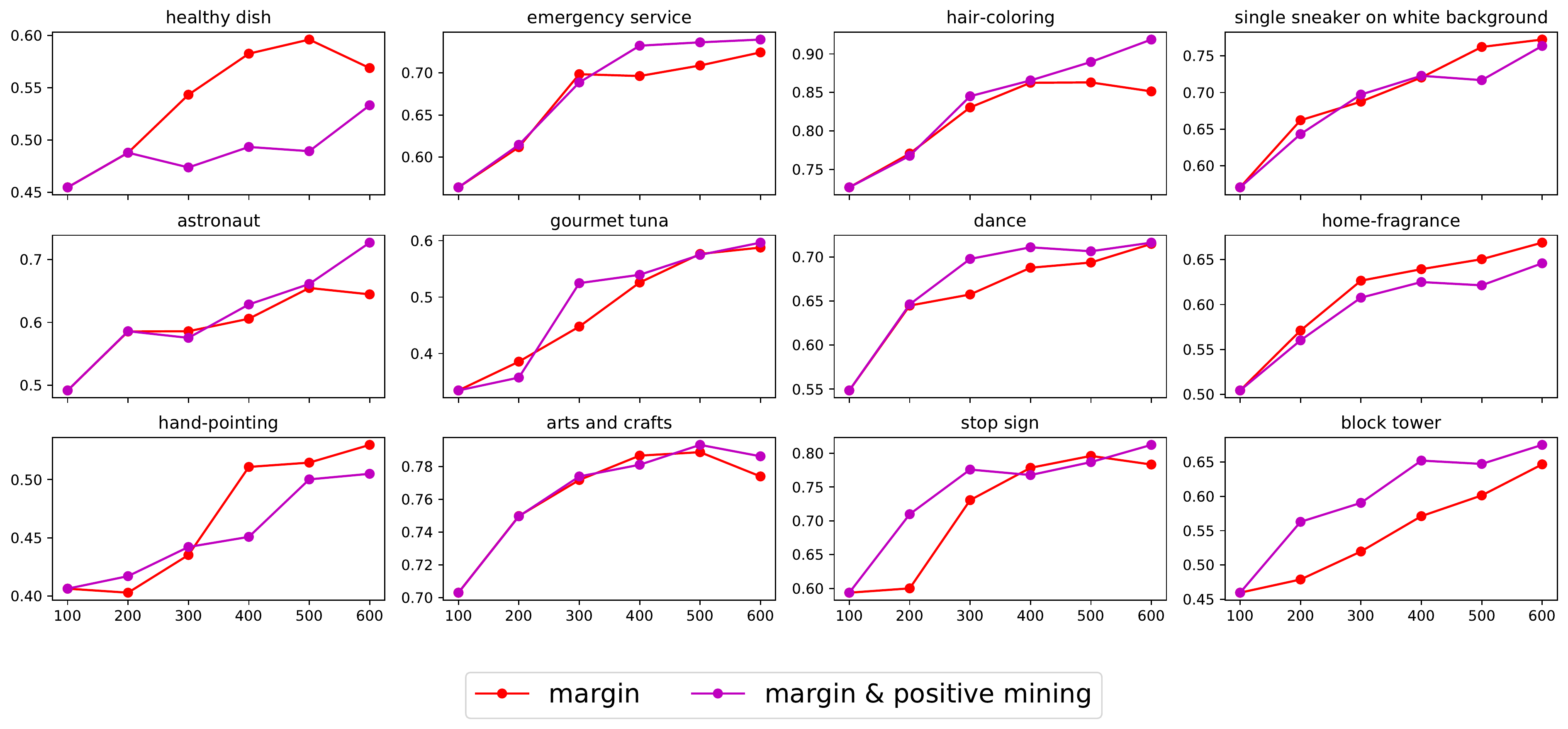}
   \caption{Results per concept for margin vs margin \& positive mining of \cite{Mullapudi}. The each figure shows the AUC PR (on y-axis) for each active learning round (on x-axis) for the two methods.}
   \label{fig:al-stanford-per-concept}
\end{figure*}

\subsection{Margin versus Margin + Positive Mining}

We show in detail the results per concept for the two active learning strategies considered in our paper: margin sampling and the margin sampling + positive mining of \cite{Mullapudi}. The results are show in Figure~\ref{fig:al-stanford-per-concept}. We observe that for the majority of the concepts the two methods are very close. Some exceptions include the concepts \concept{healthy dish} and \concept{hand pointing} for which margin performs better, while for \concept{block tower} margin + positive mining works better. Overall it is not clear that one method is significantly better than the other.

\section{Concept difficulty}
\label{appendix:concept-difficulty}

To be unbiased with respect to who the rater is---whether it is the user or crowd raters---we decided to measure concept difficulty as the performance of a zero-shot model.
We show the performance of the zero-shot model using CLIP embeddings for each concept, measured in terms of AUC PR on the test set, in Table~\ref{tab:zs-scores}.

\begin{table}[!h]
  \centering
  \small
  \begin{tabular}{lc}
    \toprule
    Concept & Score \\
    \midrule
                            gourmet tuna & 0.37 \\
                            healthy dish & 0.46 \\
                           hand-pointing & 0.47 \\
                               astronaut & 0.48 \\
                             block tower & 0.49 \\
                          home-fragrance & 0.50 \\
                               stop sign & 0.51 \\
                       emergency service & 0.53 \\
                       in-ear-headphones & 0.55 \\
      single sneaker on white background & 0.56 \\
                                   dance & 0.61 \\
                               pie-chart & 0.66 \\
                           hair-coloring & 0.73 \\
                         arts and crafts & 0.74 \\
    \bottomrule
  \end{tabular}
  \vspace{-1ex}
  \caption{Difficulty score per concept, estimated as AUC PR of the zero-shot model using CLIP embeddings.}
  \label{tab:zs-scores}
\end{table}

With these scores, we can group the top 7 easiest and top 7 hardest concepts:
\begin{itemize}
    \item top 7 easiest concepts: \concept{emergency service}, \concept{in-ear-headphones}, \concept{single sneaker on white background}, \concept{dance}, \concept{pie-chart}, \concept{hair-coloring}, \concept{arts and crafts}
    
    \item top 7 hardest concepts: \concept{gourmet tuna}, \concept{healthy dish}, \concept{hand-pointing}, \concept{astronaut}, \concept{block tower}, \concept{home-fragrance}, \concept{stop sign}
\end{itemize}

\begin{figure*}[!t]
     \centering
     \begin{subfigure}[b]{0.33\linewidth}
         \centering
         \includegraphics[width=\linewidth]{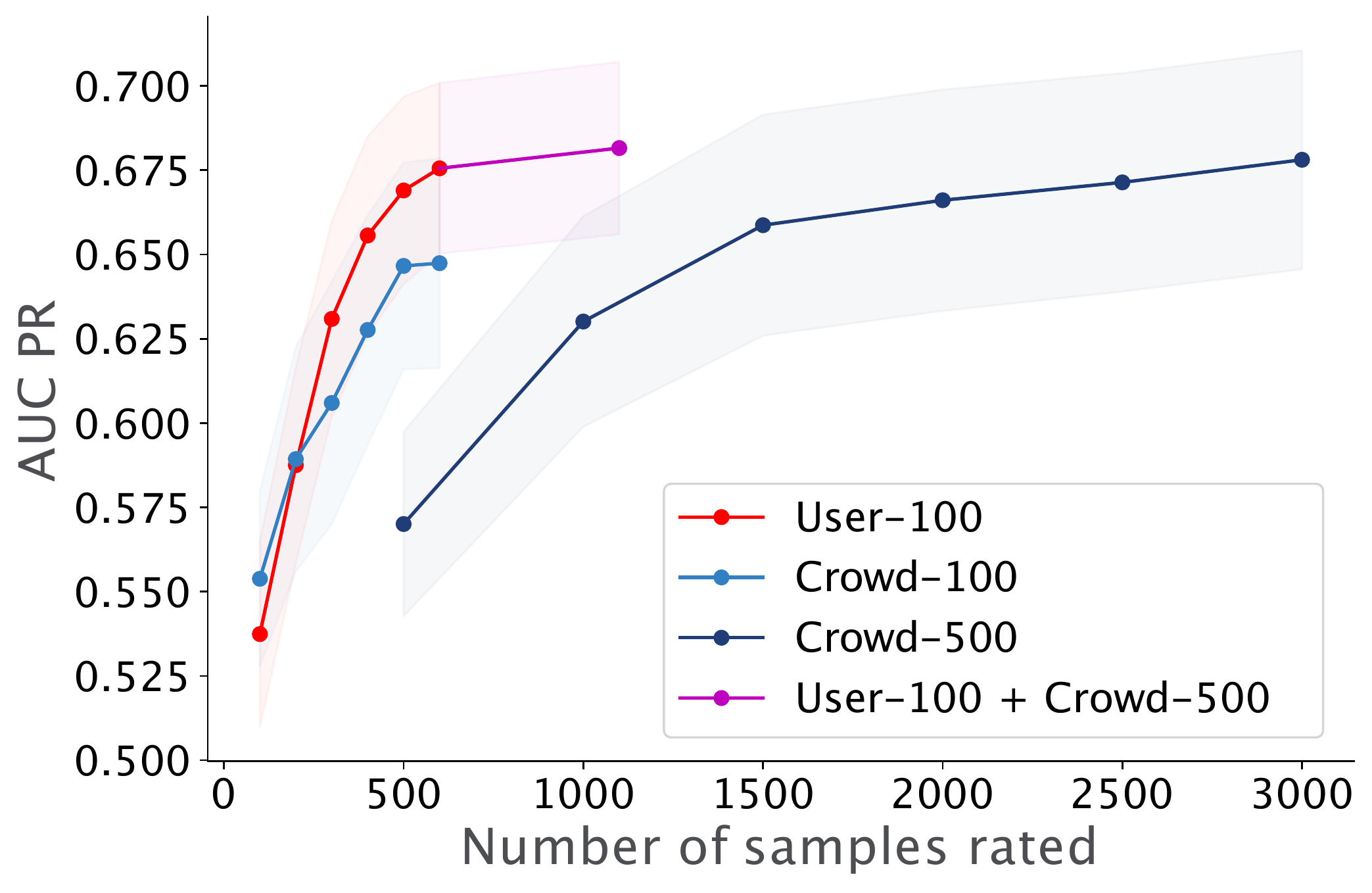}
         \caption{AUC PR.}
         \label{fig:crowd-cont-auc-pr}
     \end{subfigure}
     \hfill
     \begin{subfigure}[b]{0.33\linewidth}
         \centering
         \includegraphics[width=\linewidth]{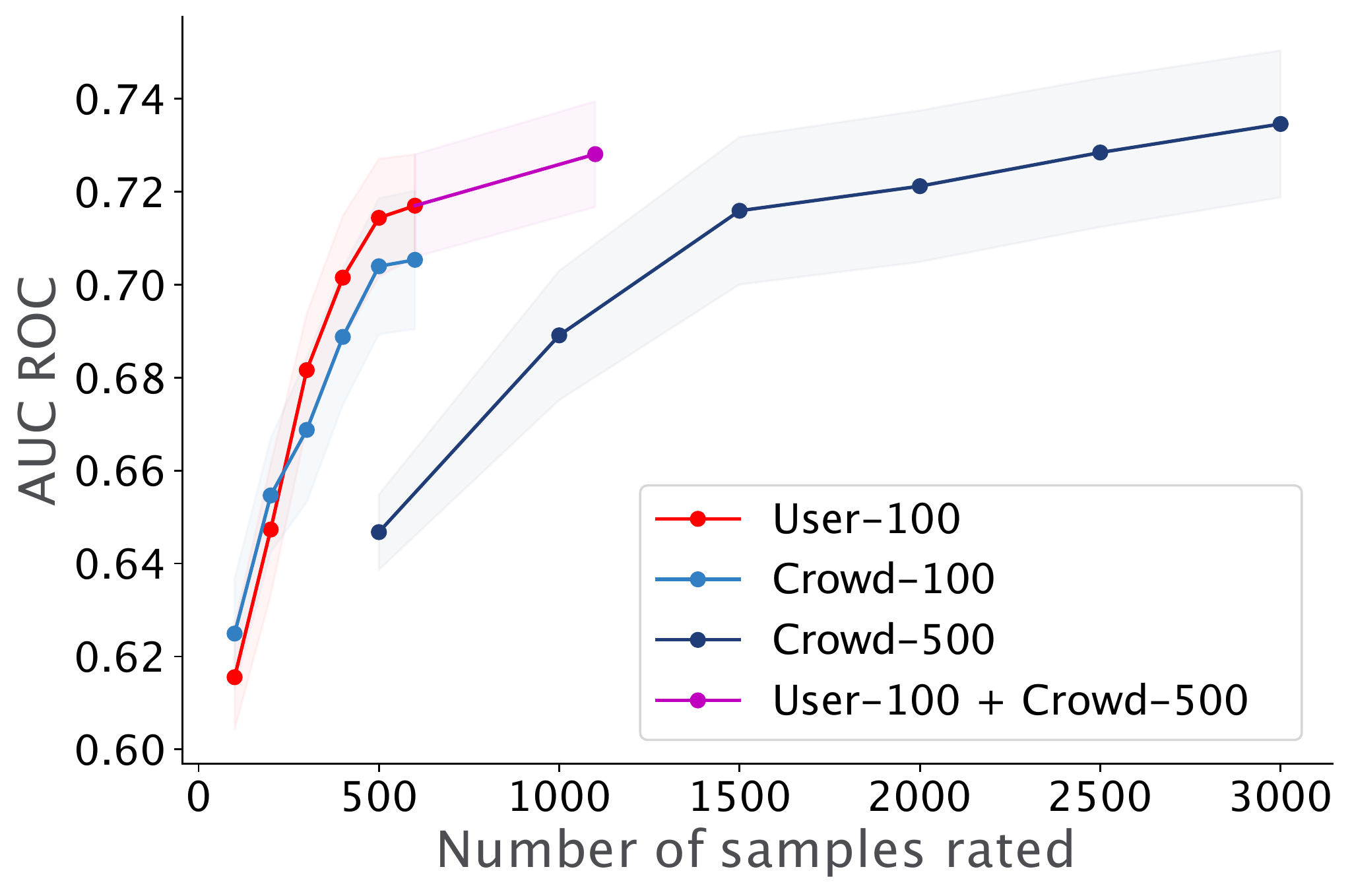}
         \caption{AUC ROC.}
         \label{fig:crowd-cont-auc-roc}
     \end{subfigure}
     \hfill
     \begin{subfigure}[b]{0.33\linewidth}
         \centering
         \includegraphics[width=\linewidth]{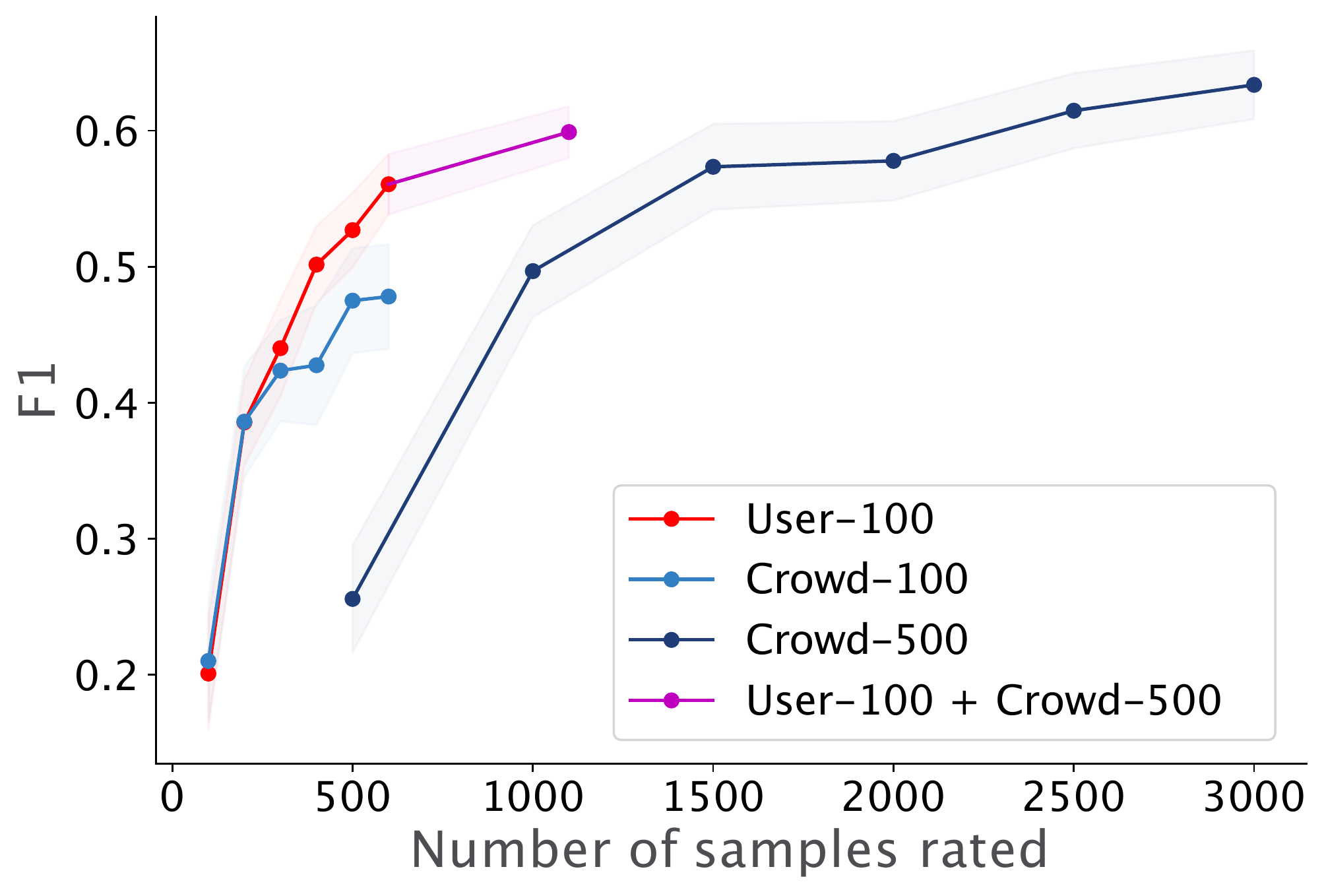}
         \caption{F1 score.}
         \label{fig:crowd-cont-f1}
     \end{subfigure}
     \caption{Model performance per amount of samples rated by the user and/or crowd raters. We also display an additional experimental setting \texttt{User-100 + Crowd-500}, where 5 rounds of user AL with batch size 100 are continued with another round of AL with crowd raters, with batch size 500.
     Mean and standard error over all concepts, for multiple metrics.}
        \label{fig:crowd-after-expert}
\end{figure*}

\section{Augmenting user labeling with crowd ratings}

One natural question to ask is what happens if we combine the benefits from doing active learning (AL) with users with those of AL with crowd raters.
We considered such a setting. For each concept, we took the model trained after 5 rounds of AL with the user (setting \texttt{User-100} in Section~\ref{sec:value-of-domain-experts}) and we used it for another round of active learning with a larger batch size (500), this time rated by crowd workers. 
The results are shown in Figure~\ref{fig:crowd-after-expert}, where we named this setting \texttt{User-100 + Crowd-500}.

With additional data from the crowd raters, the model shows further improvements.


\section{ImageNet21k experiment details}
We use these concepts in our ImageNet21k experiments:

{
\vspace{-0.5em}
\begin{center}
\textbf{50 easy concepts:}
\end{center}
\vspace{-2em}
\small
\raggedright
\begin{multicols}{2}
\begin{enumerate}
\setlength\itemsep{-0.3em}
\item tree frog \textit{ \scriptsize(n00442981)}
\item harvestman \textit{ \scriptsize(n00453935)}
\item coucal \textit{ \scriptsize(n02911485)}
\item king penguin \textit{ \scriptsize(n02955540)}
\item Irish wolfhound \textit{ \scriptsize(n02957755)}
\item komondor \textit{ \scriptsize(n02973017)}
\item German shepherd \textit{ \scriptsize(n02975212)}
\item bull mastiff \textit{ \scriptsize(n02982599)}
\item Newfoundland \textit{ \scriptsize(n02992032)}
\item white wolf \textit{ \scriptsize(n03017168)}
\item ladybug \textit{ \scriptsize(n03181293)}
\item rhinoceros beetle \textit{ \scriptsize(n03340009)}
\item leafhopper \textit{ \scriptsize(n03365991)}
\item baboon \textit{ \scriptsize(n03413828)}
\item marmoset \textit{ \scriptsize(n03439814)}
\item Madagascar cat \textit{ \scriptsize(n03454211)}
\item analog clock \textit{ \scriptsize(n03484083)}
\item apiary \textit{ \scriptsize(n03525454)}
\item bathtub \textit{ \scriptsize(n03585875)}
\item bookcase \textit{ \scriptsize(n03592245)}
\item CD player \textit{ \scriptsize(n03727837)}
\item chain mail \textit{ \scriptsize(n03779000)}
\item chest \textit{ \scriptsize(n03996145)}
\item cornet \textit{ \scriptsize(n04041544)}
\item desk \textit{ \scriptsize(n04073948)}
\item desktop computer \textit{ \scriptsize(n04236702)}
\item gondola \textit{ \scriptsize(n04288272)}
\item letter opener \textit{ \scriptsize(n04422875)}
\item microwave \textit{ \scriptsize(n04571958)}
\item nail \textit{ \scriptsize(n04586581)}
\item patio \textit{ \scriptsize(n04970916)}
\item pickup \textit{ \scriptsize(n07681926)}
\item plane \textit{ \scriptsize(n07732747)}
\item pot \textit{ \scriptsize(n07805254)}
\item purse \textit{ \scriptsize(n07815588)}
\item racket \textit{ \scriptsize(n07819480)}
\item snowplow \textit{ \scriptsize(n07820497)}
\item sombrero \textit{ \scriptsize(n07820814)}
\item stopwatch \textit{ \scriptsize(n07850083)}
\item strainer \textit{ \scriptsize(n07860988)}
\item theater curtain \textit{ \scriptsize(n07867883)}
\item ice cream \textit{ \scriptsize(n07869391)}
\item pretzel \textit{ \scriptsize(n07907161)}
\item cauliflower \textit{ \scriptsize(n07918028)}
\item acorn squash \textit{ \scriptsize(n07933891)}
\item lemon \textit{ \scriptsize(n08663860)}
\item pizza \textit{ \scriptsize(n09213565)}
\item burrito \textit{ \scriptsize(n09305031)}
\item hen-of-the-woods \textit{ \scriptsize(n13908580)}
\item ear \textit{ \scriptsize(n14899328)}
\end{enumerate}
\end{multicols}
}

{
\vspace{-0.5em}
\begin{center}
\textbf{50 hard concepts:}
\end{center}
\vspace{-2em}
\small
\raggedright
\begin{multicols}{2}
\begin{enumerate}
\setlength\itemsep{-0.3em}
\item dive \textit{ \scriptsize(n00442981)}
\item fishing \textit{ \scriptsize(n00453935)}
\item buffer \textit{ \scriptsize(n02911485)}
\item caparison \textit{ \scriptsize(n02955540)}
\item capsule \textit{ \scriptsize(n02957755)}
\item cartridge holder \textit{ \scriptsize(n02973017)}
\item case \textit{ \scriptsize(n02975212)}
\item catch \textit{ \scriptsize(n02982599)}
\item cellblock \textit{ \scriptsize(n02992032)}
\item chime \textit{ \scriptsize(n03017168)}
\item detector \textit{ \scriptsize(n03181293)}
\item filter \textit{ \scriptsize(n03340009)}
\item floor \textit{ \scriptsize(n03365991)}
\item game \textit{ \scriptsize(n03413828)}
\item glider \textit{ \scriptsize(n03439814)}
\item grapnel \textit{ \scriptsize(n03454211)}
\item handcart \textit{ \scriptsize(n03484083)}
\item holder \textit{ \scriptsize(n03525454)}
\item ironing \textit{ \scriptsize(n03585875)}
\item jail \textit{ \scriptsize(n03592245)}
\item mat \textit{ \scriptsize(n03727837)}
\item module \textit{ \scriptsize(n03779000)}
\item power saw \textit{ \scriptsize(n03996145)}
\item radio \textit{ \scriptsize(n04041544)}
\item religious residence \textit{ \scriptsize(n04073948)}
\item sleeve \textit{ \scriptsize(n04236702)}
\item spring \textit{ \scriptsize(n04288272)}
\item thermostat \textit{ \scriptsize(n04422875)}
\item weld \textit{ \scriptsize(n04571958)}
\item winder \textit{ \scriptsize(n04586581)}
\item pink \textit{ \scriptsize(n04970916)}
\item cracker \textit{ \scriptsize(n07681926)}
\item cress \textit{ \scriptsize(n07732747)}
\item mash \textit{ \scriptsize(n07805254)}
\item pepper \textit{ \scriptsize(n07815588)}
\item mustard \textit{ \scriptsize(n07819480)}
\item sage \textit{ \scriptsize(n07820497)}
\item savory \textit{ \scriptsize(n07820814)}
\item curd \textit{ \scriptsize(n07850083)}
\item dough \textit{ \scriptsize(n07860988)}
\item fondue \textit{ \scriptsize(n07867883)}
\item hash \textit{ \scriptsize(n07869391)}
\item Irish \textit{ \scriptsize(n07907161)}
\item sour \textit{ \scriptsize(n07918028)}
\item herb tea \textit{ \scriptsize(n07933891)}
\item top \textit{ \scriptsize(n08663860)}
\item bank \textit{ \scriptsize(n09213565)}
\item hollow \textit{ \scriptsize(n09305031)}
\item roulette \textit{ \scriptsize(n13908580)}
\item culture medium \textit{ \scriptsize(n14899328)}
\end{enumerate}
\end{multicols}
}
\label{appendix:imagenet}

\end{document}